\documentclass[10pt,journal,compsoc]{IEEEtran}
\ifCLASSOPTIONcompsoc
  \usepackage[nocompress]{cite}
\else
  \usepackage{cite}
\fi
\ifCLASSINFOpdf
\else
\fi
\usepackage{amsmath,amssymb,amsthm,color,bbm}
\usepackage[ruled,linesnumbered,ruled]{algorithm2e}

\usepackage{indentfirst,dblfloatfix,float,graphicx,subcaption}
\DeclareCaptionLabelSeparator{periodspace}{.\quad}
\newtheorem{assumption1}{Assumption}
\newtheorem{identity}{Identity}
\newtheorem{def1}{Definition}
\newtheorem{theorem1}{Theorem}
\newtheorem{theorem2}{Lemma}

\DeclareMathOperator*{\argmin}{arg\,min}
\DeclareMathOperator*{\argmax}{arg\,max}

\begin{document}
%
\title{Wald-Kernel: Learning to Aggregate Information for Sequential Inference}
%
%
%
%

\author{Diyan~Teng~
        and~Emre~Ertin
\IEEEcompsocitemizethanks{\IEEEcompsocthanksitem D. Teng is with the Department
of Electrical and Computer Engineering, The Ohio State University, Columbus,
OH, 43210.\protect\\
E-mail: teng.59@osu.edu
\IEEEcompsocthanksitem E. Ertin is with the Department
of Electrical and Computer Engineering, The Ohio State University, Columbus,
OH, 43210.\protect\\
E-mail: ertin.1@osu.edu
}
}

\IEEEtitleabstractindextext{%
\begin{abstract}
Sequential hypothesis testing is a desirable decision making strategy in any time sensitive scenario. Compared with fixed sample-size testing, sequential testing is capable of achieving identical probability of error requirements using less samples in average. For a binary detection problem, it is well known that for known density functions accumulating the likelihood ratio statistics is time optimal under a fixed error rate constraint. This paper considers the problem of learning a binary sequential detector from training samples when density functions are unavailable. We formulate the problem as a constrained likelihood ratio estimation which can be solved efficiently through convex optimization by imposing Reproducing Kernel Hilbert Space (RKHS) structure on the log-likelihood ratio function. In addition, we provide a computationally efficient approximated solution for large scale data set. The proposed algorithm, namely Wald-Kernel, is tested on a synthetic data set and two real world data sets, together with previous approaches for likelihood ratio estimation. Our empirical results show that the classifier trained through the proposed technique achieves smaller average sampling cost than previous approaches proposed in the literature for the same error rate.

\end{abstract}

}

\maketitle

\IEEEdisplaynontitleabstractindextext

%
\IEEEpeerreviewmaketitle

\IEEEraisesectionheading{\section{Introduction}\label{sec:introduction}}

%
%
%
%
\IEEEPARstart{S}{equential} decision strategies outperform their fixed sample-size counterparts as they can achieve any given level of decision risk using less number of samples on the average. Initially, developed by Wald~\cite{wald1947} to reduce the number of inspections in industrial quality control, sequential tests are now widely used in clinical studies to reduce the number of patients that are undergoing potentially risky treatments. Even when the cost of samples are not a major concern, sequential techniques can be used to reduce the computational cost of acquiring relevant information from a data sample. Thus sequential test is still a method of great potential in any time, sample or complexity constrained application domain. For example, in many computer vision problems, information rich features have usually high computational complexity   though they can support decisions with lower error rates. To control computational complexity of the decision rules, sequential rules such as cascading classifier such as Viola-Jones\cite{viola2001rapid} have been proposed.

For classification problems with {\em known} class conditional densities, it is well known that accumulating likelihood statistics and comparing with fixed thresholds minimizes the average stopping time under fixed error constraints. In this paper, we consider the case of {\em unknown} class conditional densities and study supervised learning of sequential decision rules from labeled data samples. There exists plethora of supervised learning algorithms to learn fixed sample test rules using parametric and non-parametric forms. In contrast,  there exists relatively few algorithms designed to learn to perform sequential classification. Unlike the single sample classification problems where only the decision boundary is critical, sequential decision rules require a mapping from sample space to a state space for aggregation of evidence and  stopping rules based on posterior confidence estimates. Here, we focus on the specific problem of learning a binary sequential classifier assume identical and conditionally independent samples from an unknown class conditional density.  For temporal aggregation of information across samples constructing an estimate of the likelihood or equivalently posterior probability emerges as an obvious framework for constructing sequential rule.

The information aggregation problem itself has been discussed in \cite{platt1999probabilistic} and the reference therein without considering a sequential decision scenario. In the same framework of this paper,  Sochman and Matas~\cite{sochman2005waldboost} constructed a likelihood ratio function estimator using Adaboost~\cite{freund1995boosting, schapire1990strength} to perform binary sequential classification based on accumulation and thresholding of the likelihood ratio estimate, resulting in an algorithm called Wald-Boost algorithm. Similarly, other methods of constructing likelihood ratio estimates based on maximizing information theoretic functionals~\cite{nguyen2010estimating, suzuki2008approximating} can be employed to perform sequential decisions. 
However, the optimization criteria used by these methods for constructing likelihood ratio functions estimates are not directly related to the performance in sequential detection. Specifically, errors in the likelihood estimate effect the average stopping time and error probabilities in a non-trivial way due to accumulation of errors across samples. Therefore, likelihood estimates tailored to optimize end-to-end performance in a sequential task are likely to outperform generic techniques.  Notably, Kuh {\em et al.}~\cite{guo1997temporal, kuh2006sequential} used reinforcement learning methods to propagate errors in terminal decisions in a sequential setting to adjust parameters of a likelihood ratio function estimate to learn binary sequential classifiers. However, again stopping time is not considered as part of the optimization criteria falling short of optimizing end-to-end performance. In this paper we derive a variational bound on the sampling cost of SPRT and associated  non-parametric log-likelihood ratio estimate which minimizes this bound as a proxy for overall sampling cost. Our empirical results show that the sequential classifier trained through the proposed technique achieves smaller average sampling cost than  learned sequential tests employing generic likelihood function estimates proposed in the literature.

The paper is organized as follows. In Section~\ref{sec:problem} we mathematically state the problem and review Wald's classic results on sequential detection. In Section~\ref{sec:wald_kernel} we describe the Wald-Kernel approach for training a sequential detector. In Section~\ref{sec:cons_comp} we evaluate statistical consistency and complexity of the Wald-Kernel algorithm. We also provide an approximation strategy for applying Wald-Kernel algorithm in large scale dataset. In Section~\ref{sec:review}, we overview relevant works on soft decision classifier. In Section~\ref{sec:empirical} we study the performance of Wald-Kernel algorithm using experiments and compare it with state-of-the-art algorithms. Finally, in Section~\ref{sec:conc} we conclude and discuss potential extensions.

\section{Problem Statement}\label{sec:problem}
In this paper, the problem of learning a binary sequential detector from training data is studied. The training data consists of $M_0$  samples  $\left\{ \mathbf{x}_1^{(0)},\mathbf{x}_2^{(0)},\cdots,\mathbf{x}_{M_0}^{(0)} \right\} $ from class 0, and $M_1$ samples $\left\{\mathbf{x}_1^{(1)},\mathbf{x}_2^{(1)},\cdots,\mathbf{x}_{M_1}^{(1)} \right\}$ from class 1, sampled i.i.d. with unknown densities $p_0(x)$ and $p_1(x)$ respectively. Each sample $\mathbf{x}_m^{(c)} \in \mathbb{R}^d$ is a $d$ dimensional feature vector from class $c \in \{0,1\}$. Throughout the paper, we assume the two density functions are defined on identical support and are distinguishable. The learning problem is to design a sequential decision making mechanism which consists of: \emph{(a)} an information aggregating rule, \emph{(b)} a stopping criterion based on the aggregated information and \emph{(c)} a decision rule when the test terminates to make a final decision between the two hypotheses $\{\mathrm{H}_0, \mathrm{H}_1\}$. Therefore, the designed classifier could be applied later on in the testing phase to classify sequentially observed data from the same source as the learning set.

\subsection{Wald's Sequential Probability Ratio Test}
In this work, we are interested in simplifying the sequential testing mechanism. We restrict our attention to information aggregating rule that is time-consistent. Meanwhile, each sample needs to be summarized minimally sufficient for later aggregation. Recall that in the classic setting for sequential detection with known class conditional density, Wald's Sequential Probability Ratio Test (SPRT) minimizes stopping time for both classes under constraints on miss detection and false alarm probability~\cite{wald1947}. SPRT compares the cumulative log-likelihood ratio up to current time slot $\mathrm{LLR}^{(T)} = \sum_{i=1}^{T} \log r(\mathbf{x}_i) = \sum_{i=1}^{T} \log \frac{p_1(\mathbf{x}_i)}{p_0(\mathbf{x}_i)}$ with fixed thresholds to choose between terminal decision or continue to sample: \begin{enumerate}
\item Declare $\mathrm{H}_0$ when $\mathrm{LLR}^{(T)} \leq a(\mathrm{P}_\mathrm{F},\mathrm{P}_\mathrm{M})$
\item Declare $\mathrm{H}_1$ when $\mathrm{LLR}^{(T)} \geq b(\mathrm{P}_\mathrm{F},\mathrm{P}_\mathrm{M})$
\item Continue to sample when $a(\mathrm{P}_\mathrm{F},\mathrm{P}_\mathrm{M}) < \mathrm{LLR}^{(T)} < b(\mathrm{P}_\mathrm{F},\mathrm{P}_\mathrm{M})$
\end{enumerate}
where $a$ and $b$ are respectively the lower and upper terminating boundaries directly related to the two types of error $\mathrm{P}_\mathrm{F}$ (\emph{a.k.a.} false alarm) and $\mathrm{P}_\mathrm{M}$ (\emph{a.k.a.} miss detection). As we may notice that in the classic SPRT, the information aggregation rule is consistent over time since the two terminating boundaries are fixed throughout the test. Also the information summarizing rule is the log-likelihood ratio function, which can be proved to be the minimal sufficient statistics in a binary classification problem. To characterize the two types of error as well as the expected test length, the following two assumptions are usually required.
\begin{assumption1}[zero-overshooting]\label{asmp1}
The maximal and minimal value of the log-likelihood ratio function are small compared with the termination boundaries $b$ and $a$ respectively:
\begin{equation}
\sup_\mathbf{x} \mathrm{LLR}(\mathbf{x}) \ll b
\end{equation}
and
\begin{equation}
\inf_\mathbf{x} \mathrm{LLR}(\mathbf{x}) \gg a
\end{equation}
\end{assumption1}
The two assumptions imply that the log-likelihood ratio value from any single sample is much smaller compared with either one the termination boundaries. Therefore, when SPRT terminate, the cumulative log-likelihood ratio equals $a$ or $b$ approximately. This assumption is also referred to as \emph{zero-overshoot} assumption. Under \emph{zero-overshoot} assumption on the cumulative log-likelihood ratio at stopping time, an approximate relationship could be sought between the terminating boundaries and the two types of error\cite[Equation 3.93, Equation 3.95]{levy}.
\begin{identity}[terminating error]\label{identity_error}
When a SPRT terminate, the two types of error could be characterized approximately using the two terminating boundaries $a$ and $b$ as:
\begin{equation}
\mathrm{P}_\mathrm{F} \approx \frac{1-e^a}{e^b-e^a}
\end{equation}
and
\begin{equation}
\mathrm{P}_\mathrm{M} \approx \frac{e^a(e^b-1)}{e^b-e^a}
\end{equation}
\end{identity}
Meanwhile, a similar relationship could be established for between the terminating boundaries and the expected stopping time~\cite[Equation 3.99, Equation 3.100]{levy}.
\begin{identity}[expected terminating time]\label{identity_time}
When a SPRT terminate, the expected number of samples observed under each hypothesis could be characterized approximately using the two terminating boundaries $a$ and $b$ as:
\begin{equation}
\mathrm{N}_0 = \mathrm{E}[N|\mathrm{H}_0] \approx \frac{b-a+ae^b-be^a}{e^b-e^a} \cdot \frac{1}{-\mathrm{D}_{01}}
\end{equation}
and
\begin{equation}
\mathrm{N}_1 = \mathrm{E}[N|\mathrm{H}_1] \approx \frac{a-b+be^{-a}-ae^{-b}}{e^{-a}-e^{-b}} \cdot \frac{1}{\mathrm{D}_{10}}
\end{equation}
where $\mathrm{D}_{01}$ and $\mathrm{D}_{10}$ represent the two Kullback-Leibler (KL) divergences between the two densities.
\end{identity}
Equivalently, one may also directly link the error with the stopping time based on Identity~\ref{identity_error} and Identity~\ref{identity_time} as:
\begin{equation}\label{sprt_samp_cost}
\begin{split}
\mathrm{N}_0 & \approx \frac{\omega_0}{\mathrm{D}_{01}} = \frac{\mathrm{P}_\mathrm{F} \log \frac{\mathrm{P}_\mathrm{F}}{1-\mathrm{P}_\mathrm{M}} + (1-\mathrm{P}_\mathrm{F}) \log \frac{1-\mathrm{P}_\mathrm{F}}{\mathrm{P}_\mathrm{M}}}{\mathrm{D}_{01}} \\
\mathrm{N}_1 & \approx \frac{\omega_1}{\mathrm{D}_{10}} = \frac{\mathrm{P}_\mathrm{M} \log \frac{\mathrm{P}_\mathrm{M}}{1-\mathrm{P}_\mathrm{F}} + (1-\mathrm{P}_\mathrm{M}) \log \frac{1-\mathrm{P}_\mathrm{M}}{\mathrm{P}_\mathrm{F}}}{\mathrm{D}_{10}} 
\end{split}
\end{equation}
Those identities above could be applied to select the terminating boundary values in the actual classification task. Typically, one would have a requirement for the two types of error (and they might be different). Therefore, the terminating boundaries could be computed based on the error constraints. Also the expected number of observations needed to achieve the constraints could be estimated (while this should not be confused with a fixed-sample test, and the test is still sequentially performed). 

\section{Wald-Kernel Likelihood Ratio Estimator}\label{sec:wald_kernel}
\begin{figure*}[t]
\hrule
\begin{equation}\label{objective_rkhs_convex}
\centering
\begin{split}
\min_{\pmb{\alpha}} & -\frac{\pi_0 \omega_0}{\frac{1}{M_0}\sum_{j=1}^{M_0} \sum_{c=1}^C \alpha_c k(\mathbf{x_j^{(0)},\mathbf{x}_c})}
+  \frac{\pi_1 \omega_1}{\frac{1}{M_1}\sum_{i=1}^{M_1} \sum_{c=1}^C \alpha_c k(\mathbf{x}_i^{(1)},\mathbf{x}_c) } + \frac{\lambda}{2} \pmb{\alpha}^\mathrm{T} \mathbf{K} \pmb{\alpha}\\
\text{s.t.} & \quad \frac{1}{M_0}\sum_{j=1}^{M_0} e^{\sum_{c=1}^C \alpha_c k(\mathbf{x_j^{(0)},\mathbf{x}_c})} \leq 1, \quad \frac{1}{M_1}\sum_{i=1}^{M_1} e^{ - \sum_{c=1}^C \alpha_c k(\mathbf{x}_i^{(1)},\mathbf{x}_c)} \leq 1
\end{split}
\end{equation}
where $\mathbf{K}$ is the Gram matrix with $\mathbf{K}(p,q) = k(\mathbf{x}_{p},\mathbf{x}_{q})$. $p,q = 1,2,\cdots,C$ are from the kernel centers.
\hrule
\begin{equation}\label{objective_rkhs_convex_gaussian}
\centering
\begin{split}
\min_{\pmb{\alpha}} & -\frac{\pi_0 \omega_0}{\frac{1}{M_0}\sum_{j=1}^{M_0} \sum_{c=1}^C \alpha_c \exp (-\frac{\|\mathbf{x}_j^{(0)} - \mathbf{x}_c\|^2}{\sigma^2})}
+  \frac{\pi_1 \omega_1}{\frac{1}{M_1}\sum_{i=1}^{M_1} \sum_{c=1}^C \alpha_c \exp (-\frac{\|\mathbf{x}_i^{(1)} - \mathbf{x}_c\|^2}{\sigma^2}) } + \frac{\lambda}{2} \pmb{\alpha}^\mathrm{T} \mathbf{K} \pmb{\alpha}\\
\text{s.t.} & \quad \frac{1}{M_0}\sum_{j=1}^{M_0} e^{\sum_{c=1}^C \alpha_c \exp (-\frac{\|\mathbf{x}_j^{(0)} - \mathbf{x}_c\|^2}{\sigma^2})} \leq 1, \quad \frac{1}{M_1}\sum_{i=1}^{M_1} e^{ - \sum_{c=1}^C \alpha_c \exp (-\frac{\|\mathbf{x}_i^{(1)} - \mathbf{x}_c\|^2}{\sigma^2})} \leq 1
\end{split}
\end{equation}
where $\mathbf{K}$ is the Gram matrix with $\mathbf{K}(p,q) = \exp (- \frac{\| \mathbf{x}_{p} - \mathbf{x}_{q}\|^2}{\sigma^2})$. $p,q = 1,2,\cdots,C$ are from the kernel centers.
\hrule
\end{figure*}

The SPRT framework is elegant not only in providing the optimal yet minimal redundant information fusion strategy but also in allowing the error-time cost being easily controlled by adjusting termination boundaries using Identity~\ref{identity_error} and Identity~\ref{identity_time}. Inspired by the structure of SPRT, an appealing choice for learning a binary sequential detector is to construct a function estimate for the log-likelihood ratio function from training samples and design termination and decision rule as threshold comparisons as in SPRT. In this section, we formulate a new algorithm for learning log-likelihood ratio function estimates that are tailored for performing sequential binary detection in the SPRT  decision structure of accumulation and thresholding. Our goal is to form a likelihood ratio estimate such that the resulting sequential decision structure minimizes the average stopping time (or equivalently the expected number of samples) for a desired level of probability of error. 

In order to obtain the proper likelihood ratio estimator described above, we derive a variational upper bound on the sampling cost of SPRT together with a pair of normalization constraints. The solution to following problem reveals a likelihood ratio function estimate linked to the sequential test performance.  

\begin{theorem1}\label{thm2}
Under a fixed error requirement, the average stopping time for SPRT can be upper bounded by the solution of the following problem:
\begin{equation}\label{objective}
\begin{split}
\min_{\hat{r} \in \mathcal{R}, \hat{r} \not\equiv 1} & -\frac{\pi_0 \omega_0}{\int  \log (\hat{r}) \mathrm{d}P_0} + \frac{\pi_1 \omega_1}{\int  \log(\hat{r}) \mathrm{d}P_1 }\\
\text{s.t.} & \int \hat{r} \mathrm{d}P_0 = 1, \quad \int \hat{r}^{-1} \mathrm{d}P_1 =1
\end{split}
\end{equation}
where $\omega_0$ and $\omega_1$ are given in~\eqref{sprt_samp_cost}, and $\mathcal{R}$ is the function space that the solution could be sought within.
\end{theorem1}
The proof is given in Appendix~\ref{proof_thm2}. As we will show in the proof, the two constraints together with the non-constant constraint guarantee that the fixed error requirement is satisfied. Meanwhile, the objective function attempts to find the likelihood ratio function estimator that better minimize the weighted testing time. The weights include the influence of class prior probability together with the error requirement. Theorem~\ref{thm2} provides a formula to tailor the likelihood ratio estimation problem to minimize the expected stopping time. In practice, the optimization problem could be solved using training data empirically. If we replace the class conditional densities with empirical distributions obtained from training data, the variational problem given in~\eqref{objective} becomes: 
\begin{equation}\label{objective_emp}
\begin{split}
\min_{\hat{r} \in \mathcal{R}, \hat{r} \not\equiv 1} & -\frac{\pi_0 \omega_0}{\frac{1}{M_0} \sum_{j=1}^{M_0} \log (\hat{r}(\mathbf{x}_j^{(0)})) } + \frac{\pi_1 \omega_1}{\frac{1}{M_1} \sum_{i=1}^{M_1} \log(\hat{r}(\mathbf{x}_i^{(1)}))}\\
\text{s.t.} & \frac{1}{M_0}\sum_{j=1}^{M_0} \hat{r}(\mathbf{x}_j^{(0)}) = 1, \quad \frac{1}{M_1}\sum_{i=1}^{M_1} \hat{r}^{-1}(\mathbf{x}_i^{(1)}) =1
\end{split}
\end{equation}

Next, we impose the RKHS structure to the log-likelihood ratio function. As a common problem in kernel method, when the training set size is big, one needs to address the tradeoff between the complexity of the optimization problem and the performance of the estimator. In this work, we suggest selecting $C$ kernel centers either through random subsampling of the training set or through k-means clustering. The former approach is simple to implement but may result in estimator to have large variance, while the later approach is heavier in computation but tends to be more stable when the k-means algorithm is properly initialized. Moreover, to avoid overfitting, one may augment the cost function \eqref{objective_emp} with an additional regularization term that penalizes the complexity of the estimated function. Finally, the equality constraints could be relaxed to inequality constraints. By Representer Theorem, we arrive at the convex problem in Equation~\eqref{objective_rkhs_convex} in general cases and Equation~\eqref{objective_rkhs_convex_gaussian} in the case of Gaussian kernel.

One may observe that since the two denominator terms are both linear functions of $\pmb{\alpha}$, convexity preserving rule for composition of functions guarantees that the objective being convex in $\pmb{\alpha}$ as long as $\pmb{\alpha}$ is properly initialized. Specifically, $\frac{1}{\cdot}$ is a convex non-increasing function for positive valued denominator and the linear function is concave, resulting in the composite function being convex when the denominator is positive. In addition, we need to guarantee that the first term in \eqref{objective_rkhs_convex} has a negative denominator while the second term has a positive denominator. This can be easily done by performing a proper initialization. Since those exponential function coefficients can be viewed as the normal of the hyperplane in terms of $\pmb\alpha$, in \eqref{objective_rkhs_convex} we need to choose the $\pmb\alpha$ vector such that it lies in the region that gives proper inner product value for both term. A natural yet simple choice of initial $\pmb\alpha$ could be the an equipartitioning vector of the two normal vectors with proper scale that meet both the constraints. Also in practice, the kernel width parameter $\sigma$ and the regularization parameter $\lambda$ can be chosen using cross validation. The resulting parameter vector $\pmb{\alpha}$ defines the estimator of the log-likelihood ratio function, which summarizes each observation into a log-likelihood that will be used in the learned SPRT. The testing phase is exactly the same as standard SPRT with known density, except that in the learned test the estimated likelihood ratio function is used as the information aggregation mapping. The resulting learned SPRT  automatically satisfies the error constraints with appropriately chosen thresholds as shown in Appendix~\ref{proof_thm_consistency2}. We summarize the Wald-Kernel algorithm in Algorithm~\ref{wald_kernel_algo1} and Algorithm~\ref{wald_kernel_algo2}. Algorithm~\ref{wald_kernel_algo1} computes all the necessary parameters required in Algorithm~\ref{wald_kernel_algo2}. And Algorithm~\ref{wald_kernel_algo2} performs the actual sequential classification task.

\begin{algorithm}[t]
\caption{Wald-Kernel Training Phase}
\label{wald_kernel_algo1}
\KwIn{Training samples $\mathbf{X}^{(0)} = \{\mathbf{x}_1^{(0)},\mathbf{x}_2^{(0)},\cdots,\mathbf{x}_{M_0}^{(0)}\}$ and $\mathbf{X}^{(1)} = \{\mathbf{x}_1^{(1)},\mathbf{x}_2^{(1)},\cdots,\mathbf{x}_{M_1}^{(1)}\}$, probability of error requirement $\mathrm{P}_\mathrm{F}$ and $\mathrm{P}_\mathrm{M}$, prior probability $\pi_0$ and $\pi_1$, choice of the reproducing kernel $k(\cdot,\cdot)$, regularization parameter $\lambda$ (optional).}
\KwOut{Kernel weights $\pmb \alpha$, kernel centers $\{\mathbf{x}_1,\mathbf{x}_2,\cdots,\mathbf{x}_C\}$, termination boundaries $a$ and $b$.}
\Begin{
Compute $a \approx \log \frac{\mathrm{P}_\mathrm{M}}{1-\mathrm{P}_\mathrm{F}}$ and $b \approx \log \frac{1-\mathrm{P}_\mathrm{M}}{\mathrm{P}_\mathrm{F}}$ by solving Equation~\eqref{error_rate1} and Equation~\eqref{error_rate2}\;
Select the kernel centers $\{\mathbf{x}_1,\mathbf{x}_2,\cdots,\mathbf{x}_C\}$ by (i) random subsampling $\mathbf{X}^{(0)}$ and $\mathbf{X}^{(1)}$ or (ii) k-means clustering on $\mathbf{X}^{(0)}$ and $\mathbf{X}^{(1)}$\;
Compute the pairwise inner product in the reproducing kernel space for all training samples with all the kernel centers $k(\mathbf{x}_j^{(0)},\mathbf{x}_c)$ and $k(\mathbf{x}_i^{(1)},\mathbf{x}_c)$ for all $j=1,2,\cdots,M_0$, $i=1,2,\cdots,M_1$ and $c=1,2,\cdots,C$, compute the Gram matrix $\mathbf{K}$ using all kernel centers $k(\mathbf{x}_p,\mathbf{x}_q)$ for $p,q=1,2,\cdots,C$\;
\SetAlgoVlined
\eIf{$\lambda$ is given}{
Solve the optimization problem~\eqref{objective_rkhs_convex}, return the optimal kernel weights $\pmb \alpha^\star$\;
}{
Choose a candidate set for $\lambda$\;
Solve the optimization problem~\eqref{objective_rkhs_convex} for each $\lambda$ candidate\;
Return the solution $\pmb \alpha^\star$ associated with the $\lambda$ that gives the lowest sampling cost estimates\;
}
}
\end{algorithm}

\begin{algorithm}[t]
\caption{Wald-Kernel Testing Phase}
\label{wald_kernel_algo2}
\KwIn{Kernel weights $\pmb \alpha$, kernel centers $\{\mathbf{x}_1,\mathbf{x}_2,\cdots,\mathbf{x}_C\}$, choice of the reproducing kernel $k(\cdot,\cdot)$, termination boundaries $a$ and $b$, prior probability $\pi_0$ and $\pi_1$, sequential testing data $\{\mathbf{x}_1,\mathbf{x}_2,\cdots\}$.}
\KwOut{Decision $\mathrm{H}_0$ or $\mathrm{H}_1$.}
initialization: set $t=0$ and $\mathrm{LLR}^{(t)} = \log \frac{\pi_1}{\pi_0}$\;
\While{$a<\mathrm{LLR}^{(t)}<b$}{
$t = t+1$\;
obtain a sample $\mathbf{x}_t$\;
compute $\log \hat{r}(\mathbf{x}_t) = \sum_{c=1}^C \alpha_c k(\mathbf{x}_t,\mathbf{x}_c)$\;
update $\mathrm{LLR}^{(t)} = \mathrm{LLR}^{(t-1)} + \log \hat{r}(\mathbf{x}_t)$\;
}
\eIf{$\mathrm{LLR}^{(t)}\geq b$}{
declare $\mathrm{H}_1$\;
}{
declare $\mathrm{H}_0$\;
}
\end{algorithm}

\section{Consistency and Complexity}\label{sec:cons_comp}

In this section, we first analyze the consistency of the proposed algorithm. We provide conditions for the Wald-Kernel solution to converge to the SPRT solution asymptotically. Then we analyze the computational complexity for solving problem~\eqref{objective_rkhs_convex_gaussian} using gradient based algorithm and discuss potential strategies for alleviating the computation when training data size is large.

\subsection{Consistency}
To make the work self contained, we first introduce the concept of entropy with bracketing~\cite[Definition 2.2]{van2000applications} for function class.
\begin{def1}[Entropy with bracketing]\label{def_entropy}
Let $\mathcal{F}$ be the class a function $f$ belongs to. Let $S_{p}(\delta,\mathcal{F},Q)$ be the smallest positive integer satisfying the following conditions:
\begin{itemize}
\item There exist pairs of functions $\{[f_j^L,f_j^U]\}_{j=1}^N$ such that $\|f_j^L-f_j^U\|_{p,Q} \leq \delta$ for any $j=1,2,\cdots,N$.
\item For any $f \in \mathcal{F}$ there exists a $j$ such that $f_j^L \leq f \leq f_j^U$.
\end{itemize} 
Then $H_{p}(\delta,\mathcal{F},Q) = \log S_{p}(\delta,\mathcal{F},Q)$ is called the $\delta$-entropy with bracketing for $L_p(Q)$ metric of $\mathcal{F}$.
\end{def1}

Next, the consistency of the Wald-Kernel estimator of likelihood-ratio function needs to be studied from two angles simultaneously. On one hand, the normalization constraints~\eqref{normalization_constraint} is forced empirically using training data. On the other hand, the objective function we minimize in~\eqref{objective_emp} also use the empirical average to approximate the integral from\eqref{objective}. Throughout the rest of this paper, we assume both KL divergence $\mathrm{D}_{01}$ and $\mathrm{D}_{10}$ are finite and the \emph{zero-overshoot} Assumption~\ref{asmp1} still holds. In addition, we require the following assumptions on the likelihood ratio function.
\begin{assumption1}[modeling loss free]\label{asmp_modeling}
Let $\mathcal{R}$ be the function space of the estimator $\hat{r}$. There exist a $\hat{r} \in \mathcal{R}$, such that:
\begin{equation}
\Pr(\mathbf{x} \in \mathcal{X} : |\hat{r}(\mathbf{x}) - r(\mathbf{x})|<\epsilon | \mathrm{H}_0) = 1,\quad \forall \text{ } \epsilon>0
\end{equation}
and
\begin{equation}
\Pr(\mathbf{x} \in \mathcal{X} : |\hat{r}(\mathbf{x}) - r(\mathbf{x})|<\epsilon | \mathrm{H}_1) = 1,\quad \forall \text{ } \epsilon>0
\end{equation}
Or equivalently, $\hat{r}(\mathbf{x}) \overset{\text{a.s.}}{\underset{P_0}{=}} r(\mathbf{x})$ and $\hat{r}(\mathbf{x}) \overset{\text{a.s.}}{\underset{P_1}{=}} r(\mathbf{x})$.
\end{assumption1}
The almost sure equality guarantees there is no modeling loss by imposing the function space restriction~\cite{nguyen2010estimating}, since this type of error is typically intractable and irrelevant to the asymptotic behavior of the estimator that we are interested in. Next, the following consistency result could be obtained.
\begin{theorem1}\label{thm_consistency1}
Suppose Assumption~\ref{asmp_modeling} is valid, and the following envelop conditions for the function space hold:
\begin{align}
& \int_{\mathbf{x}\in\mathcal{X}} \sup_{\hat{r}\in\mathcal{R}} |\log\hat{r}(\mathbf{x})| p_0(\mathbf{x}) \mathrm{d} \mathbf{x} < \infty \quad,\\
&\int_{\mathbf{x}\in\mathcal{X}} \sup_{\hat{r}\in\mathcal{R}} |\log\hat{r}(\mathbf{x})| p_1(\mathbf{x}) \mathrm{d} \mathbf{x} < \infty \quad,
\end{align}
also assume the following entropy condition for the function space $\mathcal{R}$ hold:
\begin{align}
& \lim_{m \to \infty} \frac{1}{m} \mathcal{H}_1 (\delta,\log\mathcal{R},P_{1m}) \xrightarrow{P_1} 0 \quad, \\
& \lim_{m \to \infty} \frac{1}{m} \mathcal{H}_1 (\delta,\log\mathcal{R},P_{0m}) \xrightarrow{P_0} 0 \quad.
\end{align}
where $P_{1m}$ and $P_{0m}$ are the two empirical distributions under $\mathrm{H}_1$ and $\mathrm{H}_0$ respectively. Then the solution to problem~\eqref{objective_emp} satisfies:
\begin{align}
&\lim_{m \to \infty} \int \log\hat{r} P_{1m} \xrightarrow{a.s.} \mathrm{E}[\log r |\mathrm{H}_1] \quad,\\
&\lim_{m \to \infty} \int -\log\hat{r} P_{0m} \xrightarrow{a.s.} \mathrm{E}[-\log r |\mathrm{H}_0] \quad.
\end{align}
\end{theorem1}
The proof is given in Appendix~\ref{proof_thm_consistency1}. With Theorem~\ref{thm_consistency1}, the consistency in terms of SPRT cost directly follow:
\begin{theorem1}\label{thm_consistency2}
Under the same condition in Theorem~\ref{thm_consistency1}, the probability or error cost and the sampling cost of the Wald-Kernel solution converge to those of the standard SPRT when the training set size grows:
\begin{align}
&\lim_{m \to \infty} \mathrm{P}_\mathrm{F}(\hat{r}) \xrightarrow{a.s.} \mathrm{P}_\mathrm{F} \quad,\\
&\lim_{m \to \infty} \mathrm{P}_\mathrm{M}(\hat{r}) \xrightarrow{a.s.} \mathrm{P}_\mathrm{M} \quad,\\
&\lim_{m \to \infty} \mathrm{N}_0(\hat{r}) \xrightarrow{a.s.} \mathrm{N}_0 \quad,\\
&\lim_{m \to \infty} \mathrm{N}_1(\hat{r}) \xrightarrow{a.s.} \mathrm{N}_1 \quad.
\end{align}
\end{theorem1}
The proof is given in Appendix~\ref{proof_thm_consistency2}. Hence, the Wald-Kernel solution is guaranteed to achieve standard SPRT performance when training set is large. Moreover, the asymptotic behavior requires training sample from both classes to grow to infinity to become valid.

\subsection{Complexity}\label{sec:complexity}
\begin{figure*}[t]
\hrule
\begin{equation}\label{objective_rkhs_convex_QC}
\centering
\begin{split}
\min_{\pmb{\alpha}} & -\frac{\pi_0 \omega_0}{\frac{1}{M_0}\sum_{j=1}^{M_0} \sum_{c=1}^C \alpha_c k(\mathbf{x_j^{(0)},\mathbf{x}_c})}
+  \frac{\pi_1 \omega_1}{\frac{1}{M_1}\sum_{i=1}^{M_1} \sum_{c=1}^C \alpha_c k(\mathbf{x}_i^{(1)},\mathbf{x}_c) } + \frac{\lambda}{2} \pmb{\alpha}^\mathrm{T} \mathbf{K} \pmb{\alpha}\\
\text{s.t.} & \quad \frac{1}{M_0} \sum_{j=1}^{M_0} \sum_{c=1}^C \alpha_c k(\mathbf{x}_j^{(0)},\mathbf{x}_c) + \frac{1}{2} \pmb \alpha^{\mathrm{T}} \left( \frac{1}{M_0} \sum_{j=1}^{M_0} \mathbb{K}_j^{(0)} \right) \pmb \alpha \leq 0,\\
& \quad - \frac{1}{M_1} \sum_{i=1}^{M_1} \sum_{c=1}^C \alpha_c k(\mathbf{x}_i^{(1)},\mathbf{x}_c) + \frac{1}{2} \pmb \alpha^{\mathrm{T}} \left( \frac{1}{M_1} \sum_{i=1}^{M_1} \mathbb{K}_i^{(1)} \right) \pmb \alpha \leq 0
\end{split}
\end{equation}
where $\mathbf{K}$ is the Gram matrix with $\mathbf{K}(p,q) = k(\mathbf{x}_{p},\mathbf{x}_{q})$, $\mathbb{K}_j^{(0)}(p,q) = k(\mathbf{x}_j^{(0)},\mathbf{x}_p) k(\mathbf{x}_j^{(0)},\mathbf{x}_q)$ and $\mathbb{K}_i^{(1)}(p,q) = k(\mathbf{x}_i^{(1)},\mathbf{x}_p) k(\mathbf{x}_i^{(1)},\mathbf{x}_q)$. $p,q = 1,2,\cdots,C$ are from the kernel centers.
\hrule
\end{figure*}

Consider solving problem~\eqref{objective_rkhs_convex} using a gradient based optimization procedure. For simplicity, assume the training set is balanced which means $M_1=M_0=M$. Suppose the number of kernel centers $C$ is selected $C \ll M$. We may notice that the following terms:
\begin{equation}\label{stored_terms1}
\frac{1}{M_0} \sum_{j=1}^{M_0} k(\mathbf{x}_{j}^{(0)},\mathbf{x}_c),\quad \frac{1}{M_1} \sum_{i=1}^{M_1} k(\mathbf{x}_{i}^{(1)},\mathbf{x}_c)
\end{equation}
for each $c$ as well as the Gram matrix $\mathbf{K}$ could be computed and stored beforehand. Hence, per iteration computing the gradient for the objective function has computational complexity $\mathcal{O}(C^2)$ and memory complexity $\mathcal{O}(C^2)$. To compute the gradient for the constraints in~\eqref{objective_rkhs_convex}, since the summation is for the entire exponential function which changes across iterations, the computational complexity is $\mathcal{O}(MC^2)$ and the memory complexity is $\mathcal{O}(C)$. Therefore, after combining objective with the constraints, the problem has computational complexity $\mathcal{O}(MC^2)$ and memory complexity $\mathcal{O}(C^2)$. The computation grow linearly with training data size at the rate equals squared number of kernel centers. Since the overall complexity is affected by the number of training data, in many real world examples where the dataset is extremely large, this could lead to the training time being unacceptable.

\begin{table*}[t]
\centering
\begin{tabular}{| l | c | c | c | c |}
\hline
& \multicolumn{2}{|c|}{Wald-Kernel} & \multicolumn{2}{|c|}{Wald-Kernel QC} \\\hline
& Computation & Memory & Computation & Memory\\\hline
Objective & $\mathcal{O}(C^2)$ & $\mathcal{O}(C^2)$ &$\mathcal{O}(C^2)$ &$\mathcal{O}(C^2)$\\\hline
Constraints &$\mathcal{O}(MC^2)$&$\mathcal{O}(C)$ &$\mathcal{O}(C^2)$ &$\mathcal{O}(C^2)$\\\hline
Overall &$\mathcal{O}(MC^2)$&$\mathcal{O}(C^2)$&$\mathcal{O}(C^2)$ &$\mathcal{O}(C^2)$\\\hline
\end{tabular}
\captionsetup{justification=centering}
\caption{Complexity of Wald-Kernel Algorithm}
\label{complexity_table}
\end{table*}

Here we provide one alternative solution such that the overall complexity could be fully controlled by specifying $C$ and being independent of $M$. We propose to solve an approximation of~\eqref{objective_rkhs_convex}. Since the overall computational complexity is determined currently by the constraints, approximating the constraints may lead to simpler solution. Here, we suggest a second order Taylor series approximation on each of the exponential function around 0. Therefore, we obtain the approximated problem presented in Equation~\eqref{objective_rkhs_convex_QC}. The approximation allows the parts:
\begin{equation}\label{stored_terms2}
\frac{1}{M_0} \sum_{j=1}^{M_0} \mathbb{K}_j^{(0)}, \quad \frac{1}{M_1} \sum_{i=1}^{M_1} \mathbb{K}_i^{(1)}
\end{equation}
being computed and stored beforehand together with previous terms~\eqref{stored_terms1}. Therefore, the new problem~\eqref{objective_rkhs_convex_QC} becomes a quadratically constrained (QC) program with computational complexity $\mathcal{O}(C^2)$ and memory complexity $\mathcal{O}(C^2)$. The complexity comparison between the two problems is given in Table~\ref{complexity_table}. The approximation strategy trades the space complexity for computational complexity in the constraints, resulting in a better balanced resource allocation between objective and constraints. And since the overall complexity is determined by the maximal termwise complexity, Wald-Kernel QC could potentially lead to higher scalability.

\section{Comparison to Previous Works}\label{sec:review}
In this section, we overview other approches of calculating soft decision statistics. A number of techniques have been suggested in the literature for directly estimating likelihood ratio functions which can be characterized into three classes: parametric approaches~\cite{hastie1990generalized,platt1999probabilistic}, non-parametric approaches~\cite{nguyen2010estimating, suzuki2008approximating}, and boosting approaches~\cite{sochman2005waldboost,freund1995boosting,schapire1990strength}. In principle all these methods provide probabilistic indicies that are suitable for aggregating information across multiple observations. However, the optimization criteria employed in these techniques is decoupled from the performance of these function estimates in a sequential decision task. This mismatch results in sub-optimal performance as illustrated by our empirical experiments in Section~\ref{sec:empirical}. In the subsequent sections, we select a few widely studied works for comparison.


\subsection{Parametric Approaches}\label{sec:parametric}
A basic parametric approach for calculating a probabilistic decision outcome is logistic regression developed by Cox~\cite{cox1992regression}. Logistic regression fit a linear-in-predictor model to the \emph{logit} function, which is the log-posterior ratio function:
\begin{equation}
\log \frac{p(\mathrm{H}_1|\mathbf{x})}{p(\mathrm{H}_0|\mathbf{x})} = \beta_0 + \sum_{i} \beta_i x_i
\end{equation}
The coefficients $\pmb \beta$ are typically trained by maximizing the binomial log-likelihood function using the training data:
\begin{align}
\argmax_{\pmb \beta} & \prod_{i} p(\mathrm{H}_1|\mathbf{x}_i)^{y_i} p(\mathrm{H}_0|\mathbf{x}_i)^{1-y_i}\\
\Leftrightarrow \argmax_{\pmb \beta} & \sum_{i} y_i \log p(\mathrm{H}_1|\mathbf{x}_i) + (1-y_i) p(\mathrm{H}_0|\mathbf{x}_i)
\end{align}
When the model was specified correctly, the solution has the asymptotic optimality of a maximum likelihood estimator. In~\cite{hastie1990generalized} Hastie and Tibshirani extended the basic logistic regression model to fit the \emph{logit} function using sum of smooth functions:
\begin{equation}
\log \frac{p(\mathrm{H}_1 | \mathbf{x})}{p(\mathrm{H}_0 | \mathbf{x})} = \sum_m f_m(\mathbf{x};\gamma_m) = F(\mathbf{x};\pmb {\gamma})
\label{additive_log_regression}
\end{equation}
Similarly, the parameter vector $\pmb \gamma$ could be selected through maximization of the binomial log-likelihood function. 

Later in~\cite{platt1999probabilistic}, Platt proposed to train a sigmoid function that takes the output of a Support Vector Machine (SVM) as input and convert it into a probabilistic measure. Specifically, the first step of this approach is to train a SVM classifier through standard optimization procedure:
\begin{equation}
\min_{f} \sum_{i} \left( 1 - y_i f(\mathbf{x}_i) \right)_+ + \frac{\lambda}{2} \|f\|_{\mathcal{H}}^2
\end{equation}
The second step consider taking each output value of the SVM $f(\mathbf{x}_i)$ and fit a sigmoid function for the posterior probability similar to logistic regression:
\begin{equation}
p(\mathrm{H}_1|f(\mathbf{x}_i)) = \frac{1}{1 + \exp(\beta_1 f(\mathbf{x}_i) + \beta_0)}
\end{equation}
while the sigmoid coefficients $\beta_1$ and $\beta_0$ are again trained through maximization of binomial log-likelihood function. The advantage of this two-step approach compared with previous approaches include: it could be seamlessly augmented to any existing SVM classifier to draw a probabilistic measure out of the SVM decision. Moreover, while SVM classifier is typically sparse in its structure, the same structure could be maintained for the overall soft classifier. Third, the sigmoid function in Platt's method has only two coefficients to tune, compared with the previous methods which scale with number of predictors.

Finally, for these parametric approaches reviewed in this section, the estimate for the likelihood ratio can be recovered by reverting the effect of class prior probability as:
\begin{equation}
\hat{r}(\mathbf{x}) = e^{F(\mathbf{x};\mathbf{\gamma})} \frac{p(\mathrm{H}_0)}{p(\mathrm{H}_1)}
\end{equation}
for across sample aggregation.

\subsection{Non-parametric Approaches}
In a novel direction, Nguyen~{\em et al.}~\cite{nguyen2010estimating} derived variational characterizations of $f$-divergences which enabled estimation of divergence functionals and likelihood ratios through convex risk minimization. They view the divergence quantity:
\begin{equation}
\mathrm{D}(p_1 \| p_0) = \int \phi \left( \frac{p_0}{p_1} \right) \mathrm{d}P_1
\end{equation}
as convex functional of the likelihood ratio function $\frac{p_0}{p_1}$, where $\phi(\cdot) = -\log(\cdot)$ in the case of KL divergence. Next, they apply Legendre–Fenchel transformation to obtain an maximization problem, whose solution serves as a lower bound to the KL divergence:
\begin{equation}
\int -\log \left( \frac{p_0}{p_1} \right) \mathrm{d}P_1 \geq \sup_{f} \int \left[ f \mathrm{d}P_0 + \mathrm{d}P_1 + \log(-f) \mathrm{d}P_1 \right]
\end{equation}
Finally, they solve the optimization problem by imposing RKHS structure on either $f$ or $\log f$ to obtain a convex problem. And they empirically replace the integration by summation using the training data:
\begin{equation}\label{kldf}
\sup_{f \in \mathcal{H}} \frac{1}{M_0} \sum_{j} f(\mathbf{x}_j^{(0)}) + \frac{1}{M_1} \sum_{i} \log(-f(\mathbf{x}_i^{(1)}))
\end{equation}
Later in~\cite{kanamori2009least,kanamori2012statistical}, Kanamori~{\em et al.} extended the $f$-divergences estimation framework to consider a least-square version of the problem. Specifically, the least-square version of the divergence functional choose:
\begin{equation}
\phi \left( \frac{p_0}{p_1} \right) = \frac{1}{2} \left\lvert\frac{p_0}{p_1} - 1 \right\rvert^2
\end{equation}
This choice resulting in a quadratic program when RKHS restriction is imposed on $\frac{p_0}{p_1}$:
\begin{equation}\label{ulsif}
\sup_{f \in \mathcal{H}} -\frac{1}{2M_1} \sum_{i} (f(\mathbf{x}_i^{(1)}))^2 + \frac{1}{M_0} \sum_{j} f(\mathbf{x}_j^{(0)})
\end{equation}
Typically, for both~\eqref{kldf} and~\eqref{ulsif}, a penalization for the RKHS norm $\|f\|_\mathcal{H}$ needs to be introduced to avoid overfitting. 

Under this non-parametric framework, the likelihood ratio function becomes a byproduct while the main objective is to estimate the f-divergence quantity value. The authors in~\cite{nguyen2010estimating} also studied the closeness between the estimated likelihood ratio and the true likelihood ratio in terms of a generalized Hellinger metric. Even though the theoretical result in~\cite{nguyen2010estimating} is convincing, directly applying the likelihood ratio estimator from this approach to any classification tasks could result in poor performance, especially in practical problems where number of training data is limited. This will be demonstrated in Section~\ref{sec:empirical} through real examples.

\subsection{Boosting Approaches}
Classic parametric approaches reviewed in Section~\ref{sec:parametric} attempt to search for the optimal model parameters in a single optimization problem. In contrast boosting, an alternative approach proposed originally by Freund and Schapire~\cite{schapire1990strength,freund1995boosting}, allows constructing the discriminative function iteratively. They propose to add a weak learner to the function in each iteration whose output value will be weighted by a coefficient related to its classification error on training samples with simultaneously updated weights. Specifically, each training sample is initialized with an equal weight. In each iteration, the weak classifier is selected according to:
\begin{equation}
\argmin_{f_i} \sum_{n} w_n^{(i)} \mathbbm{1} \left( f_i(\mathbf{x}_n) \neq y_n \right)
\end{equation}
where $w_n^{(i)}$ is the weight for the $n$th sample at the $i$th iteration, $\mathbbm{1}(\cdot)$ is the indicator function which takes value 1 when the classification result mismatches the label. Next, to calculate the weight associated with the $i$th weak classifier, they evaluate a weighted score:
\begin{equation}
\epsilon_i = \frac{\sum_{n} w_n^{(i)} \mathbbm{1} \left( f_i(\mathbf{x}_n) \neq y_n \right) }{ \sum_{n} w_n^{(i)} }
\end{equation}
which relates to the weight of weak classifier through:
\begin{equation}
c_i = \log \frac{1-\epsilon_i}{\epsilon_i}
\end{equation}
Finally, the weight for each sample in the next iteration is updated according to:
\begin{equation}
w_n^{(i+1)} = w_n^{(i)} \exp \left( c_i \mathbbm{1} \left( f_i(\mathbf{x}_n) \neq y_n \right) \right)
\end{equation}
The resulting combined classifier takes the following form:
\begin{equation}
F_A(\mathbf{x}) = \text{sign} \left( \sum_i c_i f_i(\mathbf{x}) \right)
\end{equation}

Interestingly, as shown by Friedman~{\em et al.}~\cite{friedman2000additive} boosting approaches that combine binary decision of weak classifiers to train classifiers with improved performance can be analyzed under the same framework of fitting an additive model through maximization of likelihood. They show that the iterative weighted minimization procedure is equivalent to minimizing the expected exponential error $\mathrm{E}(e^{-y F(\mathbf{x})})$, which is a second order approximation to the binomial log-likelihood function. And they pointed out that the likelihood ratio can be retrieved from the final classifier through:
\begin{equation}
\hat{r}(\mathbf{x}) = e^{2 F_A(\mathbf{x})} \frac{p(\mathrm{H}_0)}{p(\mathrm{H}_1)}
\label{adaboost}
\end{equation}
The boosting approach is fast with good empirical performance and is resistant to over-fitting and provides a fast approach for constructing likelihood ratio estimates. Therefore, Sochman and Matas proposed Wald-Boost~\cite{sochman2005waldboost} algorithm that apply AdaBoost method to learn the likelihood ratio function and combine it with Wald's SPRT framework for performing a sequential test.

Boosting based approaches have a main drawback when they are applied in soft decision making process. The original purpose of boosting is to alleviate the difficulty in designing the classifier in a single step by iteratively train many classifiers of much simpler form. Therefore, even though the theoretical analysis of the algorithm guarantees that the combined classifier approaches the optimal single-shot decision boundary asymptotically, for inputs that are far from the decision boundary, it usually provides poor estimates of the likelihood ratio value due to the oversimplification of the individual weak classifier. In other words, the procedure overemphasize the likelihood ratio area that are close to 1 and allocate all the modeling resource for those area.

\section{Experimental Results}\label{sec:empirical}
In this section, we simulate the proposed Wald-Kernel algorithm on one synthetic example as well as two real world examples. In the synthetic example, we first demonstrate the testing performance difference between the sequential classifier obtained through the original Wald-Kernel exact formula~\eqref{objective_rkhs_convex} and the one obtained through the quadratic constrained version~\eqref{objective_rkhs_convex_QC}. Meanwhile, we compare Wald-Kernel QC version with state-of-the-art algorithms for likelihood ratio estimation. We analyze the benefits of Wald-Kernel algorithm through empirically calculated histogram of the log-likelihood ratio values, together with ground truth obtained by plugging observations into class conditional density function ratio values. Next, we simulate Wald-Kernel algorithm on two real world examples. Empirical results indicate that Wald-Kernel outperforms other likelihood ratio estimation approaches. Matlab implementation of Wald-Kernel QC version is available to download at: ({https://osu.box.com/v/Wald-Kernel-demo}).

\subsection{Synthetic Example}
\begin{figure}[h]
\centering
\includegraphics[width=\linewidth]{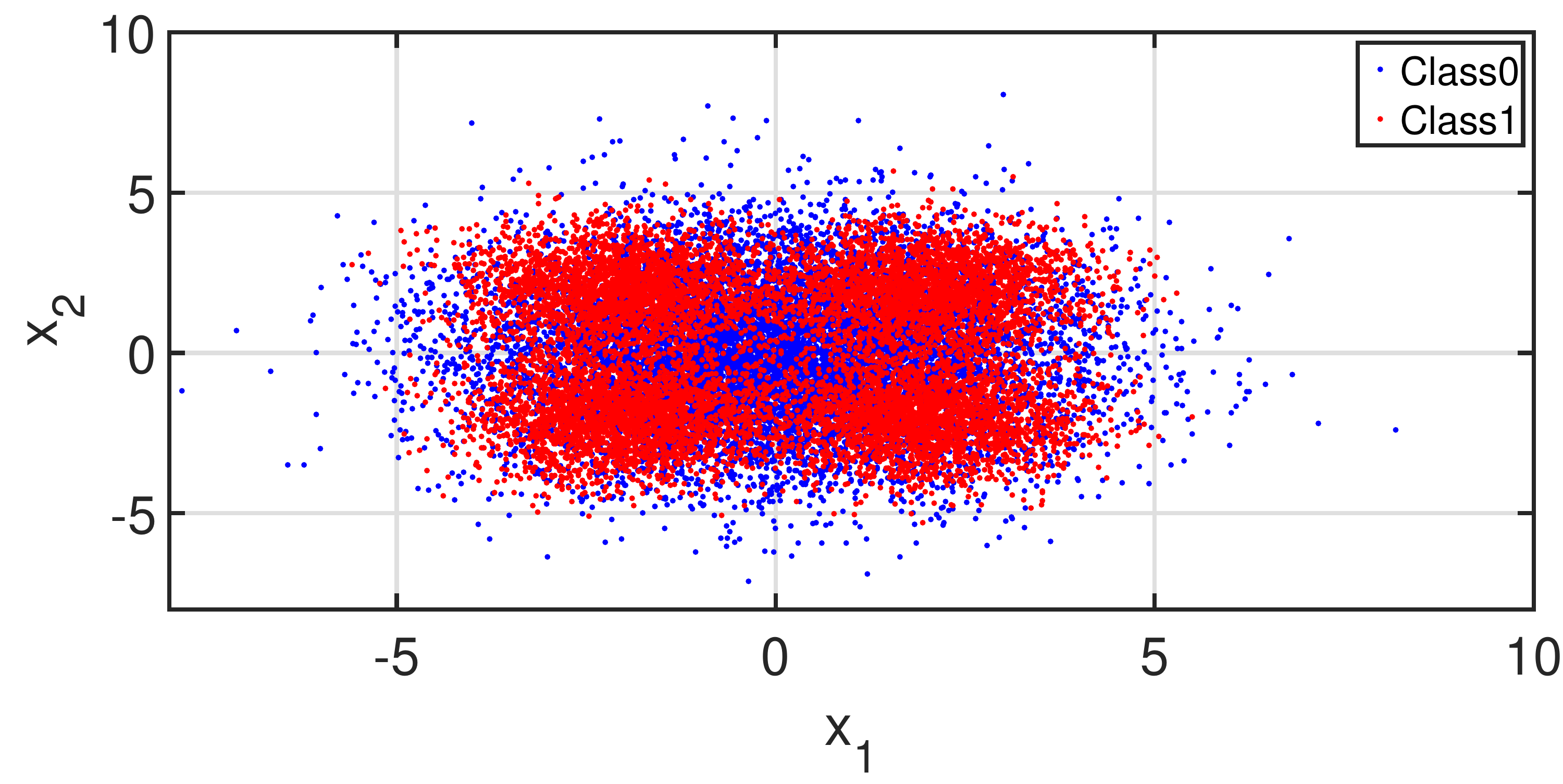}
\captionsetup{justification=centering}
\caption{Synthetic dataset visualization}
\label{synthetic_data}
\end{figure}

We first provide a synthetic example. In this example, the observation under each class is generated according to:
\begin{align*}
\mathrm{H}_0 : \quad \mathbf{x} \sim & \mathcal{N}( 
\begin{bmatrix}
0\\
0
\end{bmatrix},
\begin{bmatrix}
4&0\\
0&4
\end{bmatrix}
)\\
\mathrm{H}_1 : \quad \mathbf{x} \sim & \frac{1}{4}\mathcal{N}( 
\begin{bmatrix}
2\\
2
\end{bmatrix},
\begin{bmatrix}
1&0\\
0&1
\end{bmatrix}
)
+
\frac{1}{4}\mathcal{N}( 
\begin{bmatrix}
-2\\
2
\end{bmatrix},
\begin{bmatrix}
1&0\\
0&1
\end{bmatrix}
)\\
+ &
\frac{1}{4}\mathcal{N}( 
\begin{bmatrix}
-2\\
-2
\end{bmatrix},
\begin{bmatrix}
1&0\\
0&1
\end{bmatrix}
)
+
\frac{1}{4}\mathcal{N}( 
\begin{bmatrix}
2\\
-2
\end{bmatrix},
\begin{bmatrix}
1&0\\
0&1
\end{bmatrix}
)
\end{align*}
where we have a 2D unimodal Gaussian distribution under $\mathrm{H}_0$ and a Gaussian mixture distribution under $\mathrm{H}_1$. We generate 10,000 samples for each class as the training data which are plotted in Figure~\ref{synthetic_data} for visualization. For testing, we assume the prior probability $\pi_0 = \pi_1 = 0.5$ are known and accurate. Also, throughout this example, we choose the probability of error requirements $\mathrm{P}_\mathrm{F} = \mathrm{P}_\mathrm{M}$ for simplicity. This simplification results in the weights $\omega_0$ and $\omega_1$ being equal regardless of the actual value of $\mathrm{P}_\mathrm{F}$ and $\mathrm{P}_\mathrm{M}$.

\subsubsection{Effect of Quadratic Constraints Approximation}
As we explained in Section~\ref{sec:complexity}, Wald-Kernel QC version~\eqref{objective_rkhs_convex_QC} has higher scalability compared with standard Wald-Kernel~\eqref{objective_rkhs_convex}. Thus, for all the examples in this section, we are implementing the Wald-Kernel QC version when we are comparing with other approaches. Nevertheless, the effect of constraints approximation has not been evaluated quantitatively. Thus, here we evaluate the performance difference between Wald-Kernel QC version~\eqref{objective_rkhs_convex_QC} and the Wald-Kernel exact version~\eqref{objective_rkhs_convex}. We start with training the Wald-Kernel QC model. First, we select 200 kernel centers using k-means clustering algorithm on the 20,000 class mixed training set. Next, since we chose Gaussian kernel, we create a candidates pool for the kernel width parameter $\sigma$ and a candidates pool for the regularizer $\lambda$. For $\sigma$ pool we first compute all pairwise Euclidean distance between the 200 kernel centers and all training samples. Then we picked the $\{1,10,20,\cdots,90\}$th percentiles of the pairwise distance as the $\sigma$ pool. For the $\lambda$ pool, we chose $10^{ \{-9,-7,\cdots,-1\}}$. Next, we optimize the kernel weights $\pmb \alpha$ for each pair of the $\sigma$, $\lambda$ choice using a 50/50 cross-validation method. The final solution is selected to be the pair that provides the lowest objective function value in the validation set. After training the Wald-Kernel QC model, we input the same kernel centers, $\sigma$, and $\lambda$ value, together with the Wald-Kernel QC minimizer $\pmb \alpha_{\mathrm{QC}}^{\star}$ to the Wald-Kernel exact problem for further optimization. Finally, we obtain the Wald-Kernel exact minimizer $\pmb \alpha_{\mathrm{EXACT}}^{\star}$.

\begin{figure*}[h]
\begin{subfigure}[t]{0.33\linewidth}
\includegraphics[width=\textwidth]{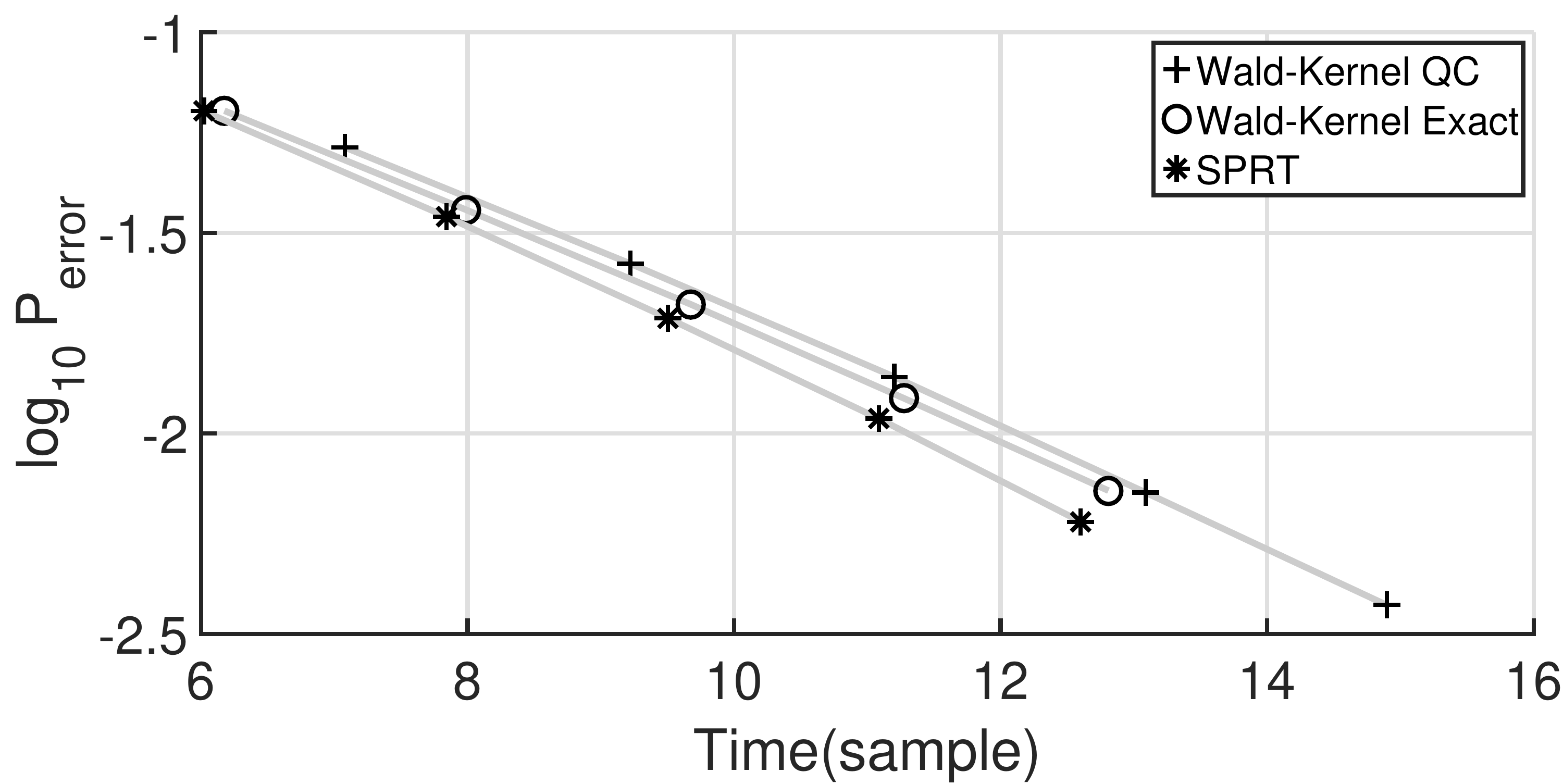}%
\caption{Class mixed probability of error \emph{vs.} testing time}%
\label{QC_VS_EXACT_err_time}
\end{subfigure}\hfill
\begin{subfigure}[t]{0.33\linewidth}
\includegraphics[width=\textwidth]{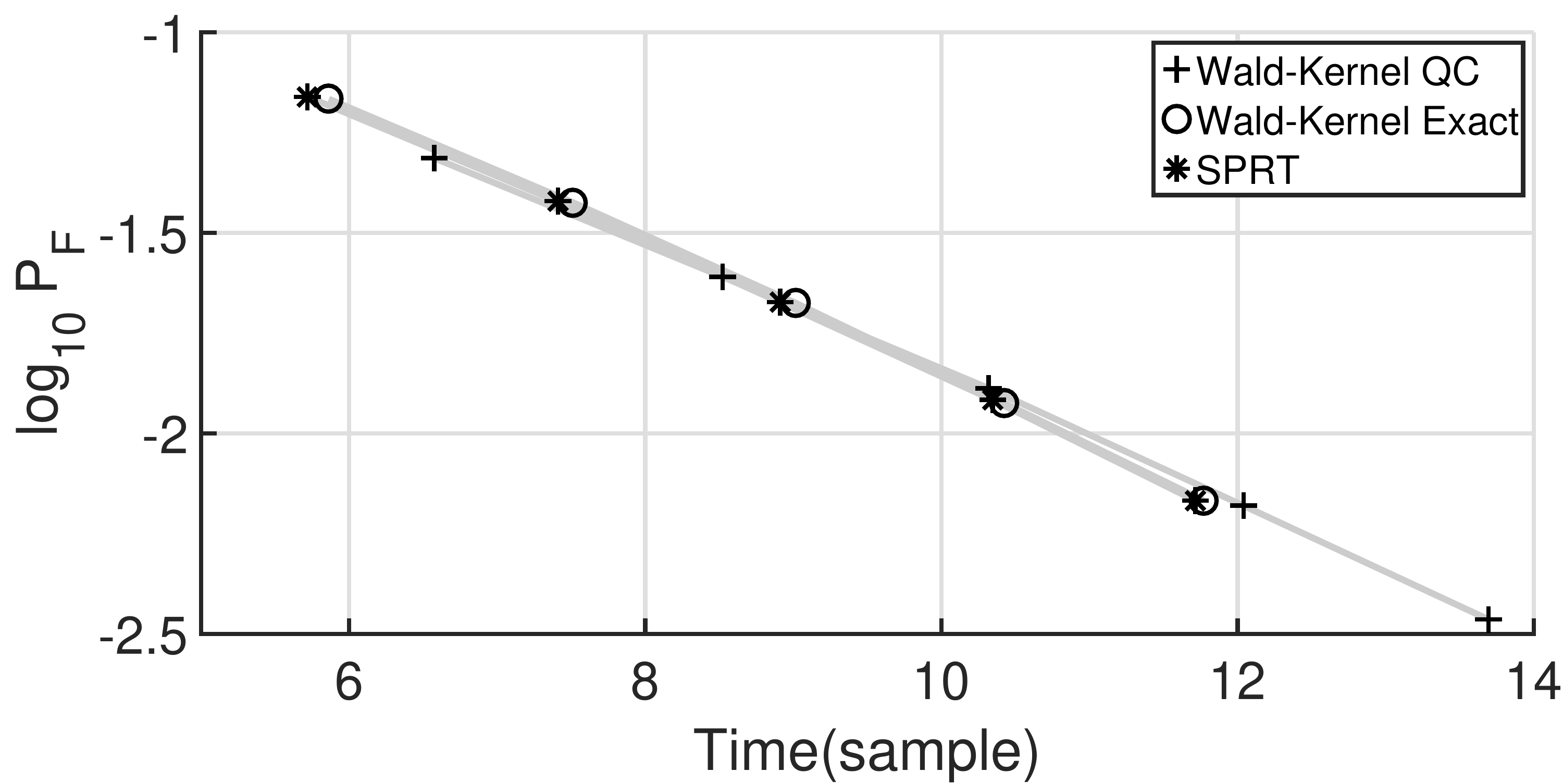}%
\caption{Probability of false alarm \emph{vs.} testing time}%
\label{QC_VS_EXACT_pf_time}
\end{subfigure}\hfill
\begin{subfigure}[t]{0.33\linewidth}
\includegraphics[width=\textwidth]{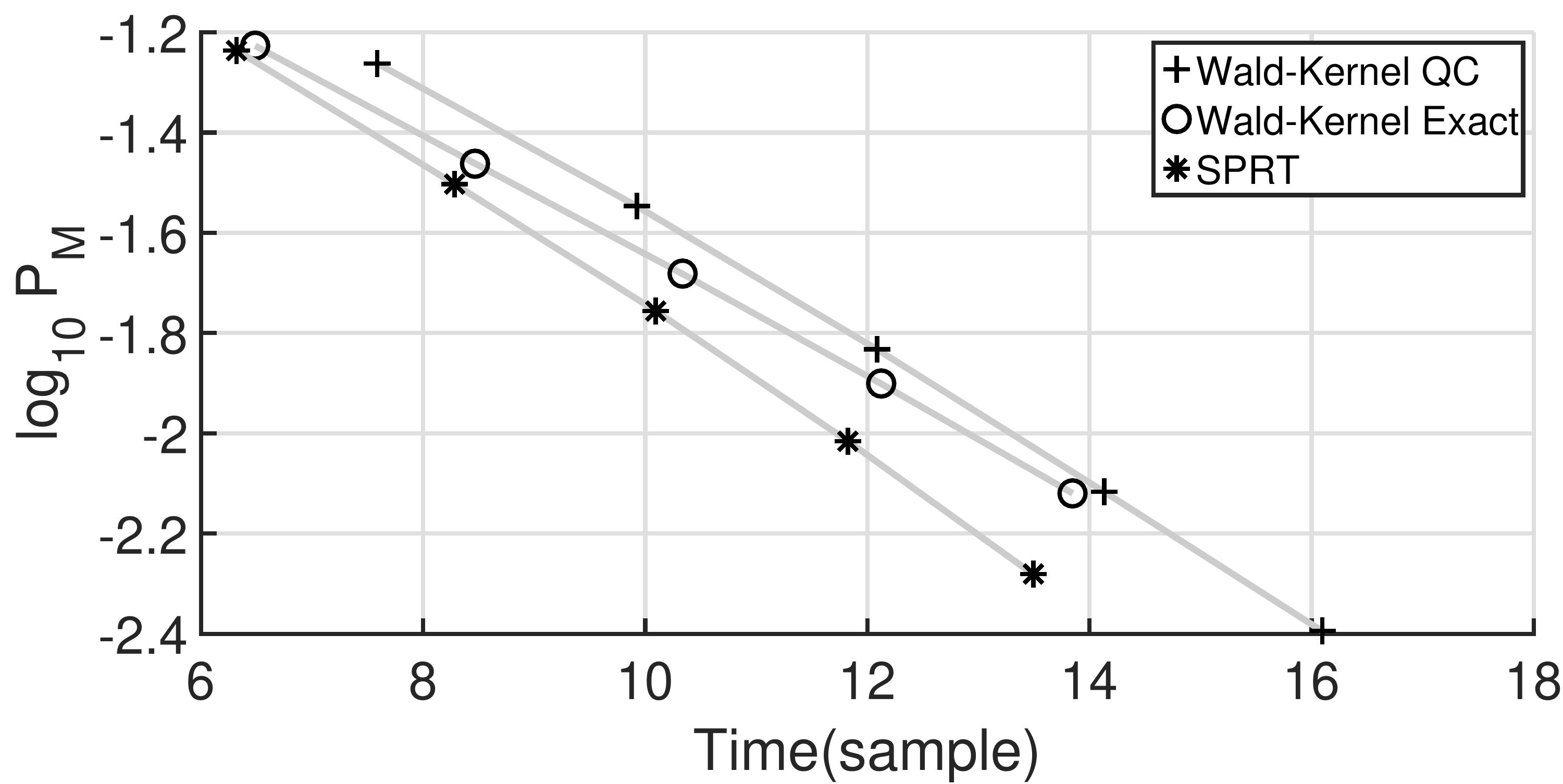}%
\caption{Probability of miss detection \emph{vs.} testing time}%
\label{QC_VS_EXACT_pm_time}
\end{subfigure}
\captionsetup{justification=centering}
\caption{Wald-Kernel QC \emph{vs.} Exact: probability of error \emph{vs.} testing time under each class}
\label{QC_VS_EXACT_err_time_performance}
\hrule
\end{figure*}

\begin{figure*}[h]
\centering
\includegraphics[width=\linewidth]{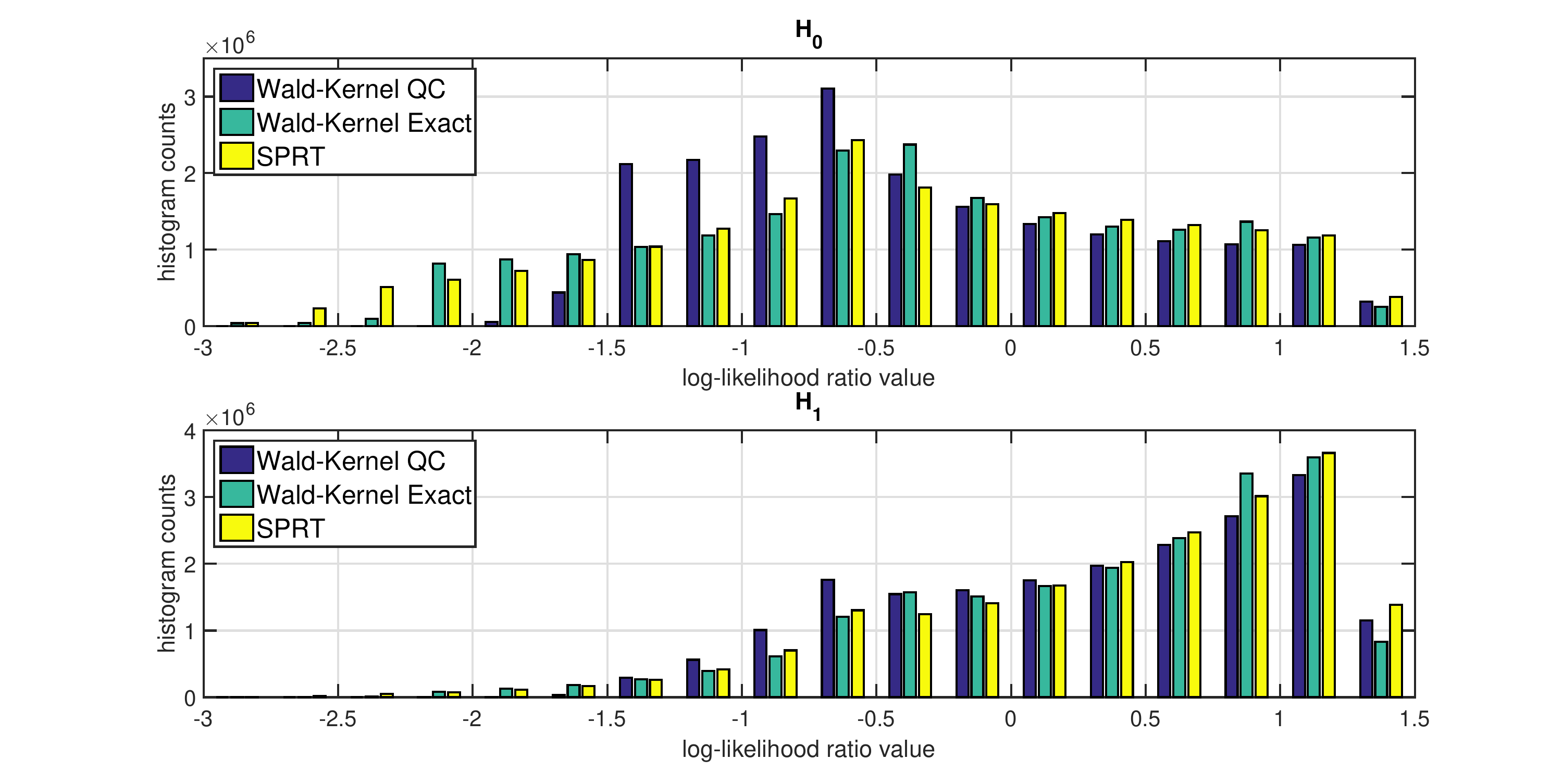}%
\captionsetup{justification=centering}
\caption{Wald-Kernel QC \emph{vs.} Exact: histogram of estimated log-likelihood ratio value under each class}%
\label{QC_VS_EXACT_histogram}
\hrule
\end{figure*}

After obtaining the optimized model, the testing phase is completed by sequentially generate and classify on 400,000 realization runs according to mechanism in Algorithm~\ref{wald_kernel_algo2}. Here the probability of error requirements are chosen across $10^{\{-1 ,-1.25,-1.5,-1.75,-2\}}$. The identical realizations are also fed into Wald's SPRT which knows the two class conditional density functions for comparison purpose. Not to be confused that the Wald's SPRT test is the optimal classifier when density functions are known. It cannot be surpassed by any learned classifier. We plotted the testing results in Figure~\ref{QC_VS_EXACT_err_time_performance}. The error rate is obtained by dividing the total number of mis-classified samples by the number of realization runs. And the termination time is obtained by dividing the recorded number of samples to terminate by the total number of realization runs. For the 5 different probability of error requirements, we compute the termination boundaries $a$ and $b$ according to the approximation formula in Algorithm~\eqref{wald_kernel_algo1}. As we may observe in Figure~\ref{QC_VS_EXACT_pf_time}, the overall performance of Wald-Kernel exact solution requires less sample to terminate under the same probability of error requirement compared with Wald-Kernel QC. Moreover, for the same termination boundaries, Wald-Kernel exact performance is closer to the SPRT solution, while Wald-Kernel QC tends to achieve a lower error rate but requires more samples to terminate. The explanation is that the normalization of likelihood ratio function is approximated in the QC version. This results in inaccurate scaling on the likelihood ratio function. Nevertheless, other than this inaccurate scaling, the overall performance degradation seems acceptable. Next, we evaluate the performance separately for each class. In Figure~\ref{QC_VS_EXACT_pf_time}, the testing performance difference under $\mathrm{H}_0$ is almost negligible. In comparison, in Figure~\ref{QC_VS_EXACT_pm_time}, the Wald-Kernel QC solution is slightly worse than the Wald-Kernel exact solution, resulting in the overall class mixed performance being negligibly worse.

Next, we plotted the computed log-likelihood ratio values for all the samples fed into the classifier during the 400,000 realization runs as a histogram format in Figure~\ref{QC_VS_EXACT_histogram}. The histogram is created by grouping log-likelihood ratio values within bins separated as $\{-3:0.25:1.5\}$. We may observe that the Wald-Kernel QC solution tends to predict log-likelihood ratio values \emph{conservatively} (which means closer to 0), while Wald-Kernel exact solution tends to match the ground truth better. Also, the Wald-Kernel QC solution claims more samples within the region $(-1.5,-0.5)$
under both hypotheses compared with the Wald-Kernel exact solution. This also explains why the miss detection rate being higher for Wald-Kernel QC, since the classifier tends overestimate the $\mathrm{H}_0$ case.

Finally, we may conclude that Wald-Kernel QC seems acceptable in terms of information fusing accuracy. Since the normalization approximation is causing the estimator to become more conservative, even though computationally more expensive one may choose the Wald-Kernel exact solution in cases where error rate requirements need to be guaranteed with minimum testing time (in other words, error rate lower than the requirement is unnecessary). In any other cases where time-accuracy tradeoff is addressed interchangeably (this means there exists a trading parameter between unit error rate decrement and unit time increment), Wald-Kernel QC could be selected.

\subsubsection{Performance Comparison Using Synthetic Dataset}
Next, we compare the performance of Wald-Kernel with other state-of-the art approaches for likelihood ratio estimation on the same synthetic dataset. These approaches includes Wald-Boost~\cite{sochman2005waldboost}, Probabilistic Valued Support Vector Machine~\cite{platt1999probabilistic}, and the one-sided KL divergence estimator~\cite{nguyen2010estimating}. For Wald-Boost, the original Ada-Boost weak-learner is chosen as a two-layer decision tree with total number of weak-learners being 200.  For Probabilistic SVM, we use Gaussian kernel SVM with kernel width $\sigma_{\mathrm{SVM}}$ 10-fold cross-validated together with the posterior probability transformation coefficients on the same cross validation sets. For one-sided KL divergence estimator, we choose to estimate $\mathrm{D}_{10}$ using identical kernel centers and width as Wald-Kernel's choice. 

\begin{figure*}[h]
\begin{subfigure}[t]{0.33\linewidth}
\includegraphics[width=\textwidth]{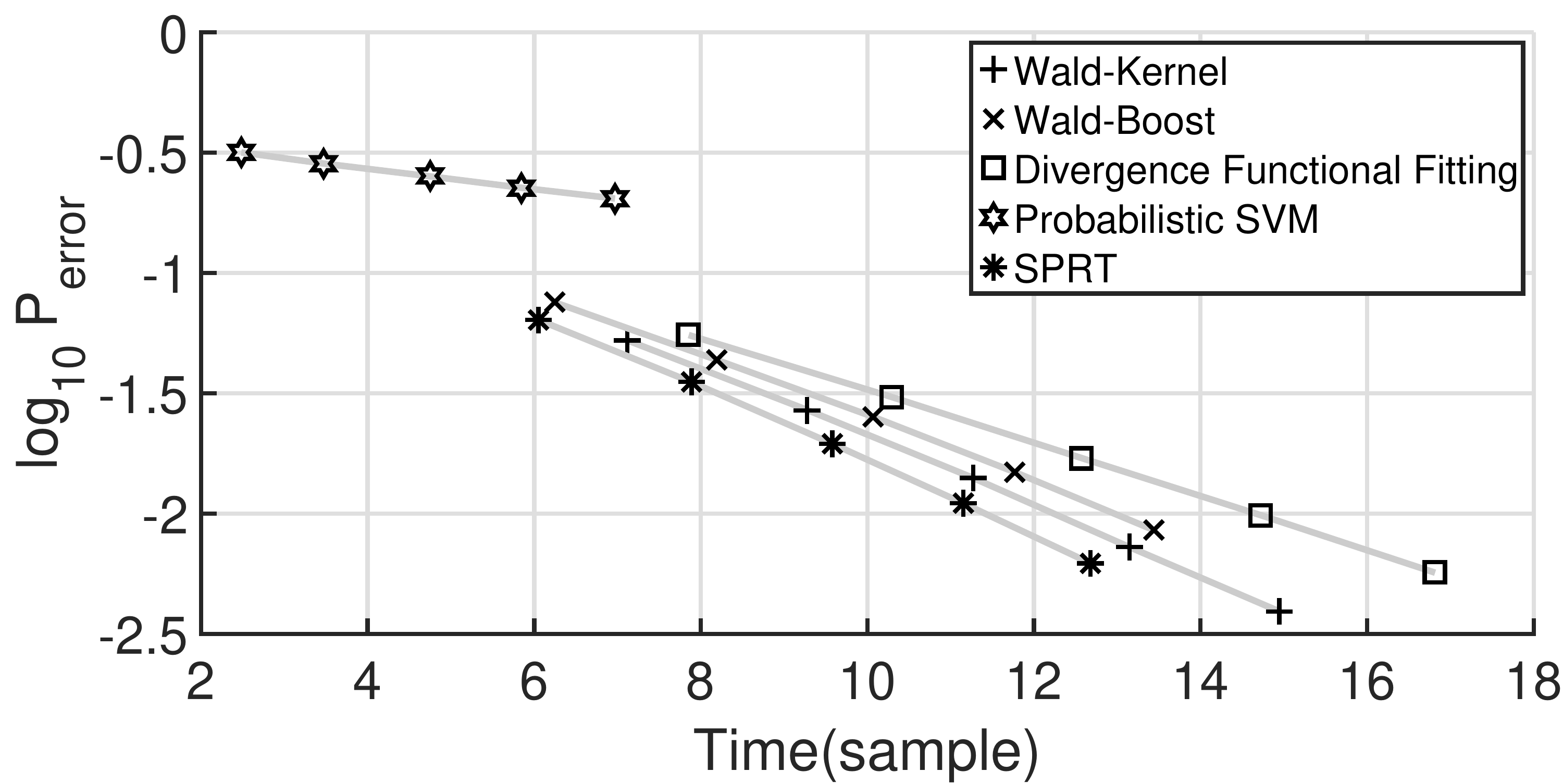}%
\caption{Class mixed probability of error \emph{vs.} testing time}%
\label{synthetic_err_time}
\end{subfigure}\hfill
\begin{subfigure}[t]{0.33\linewidth}
\includegraphics[width=\textwidth]{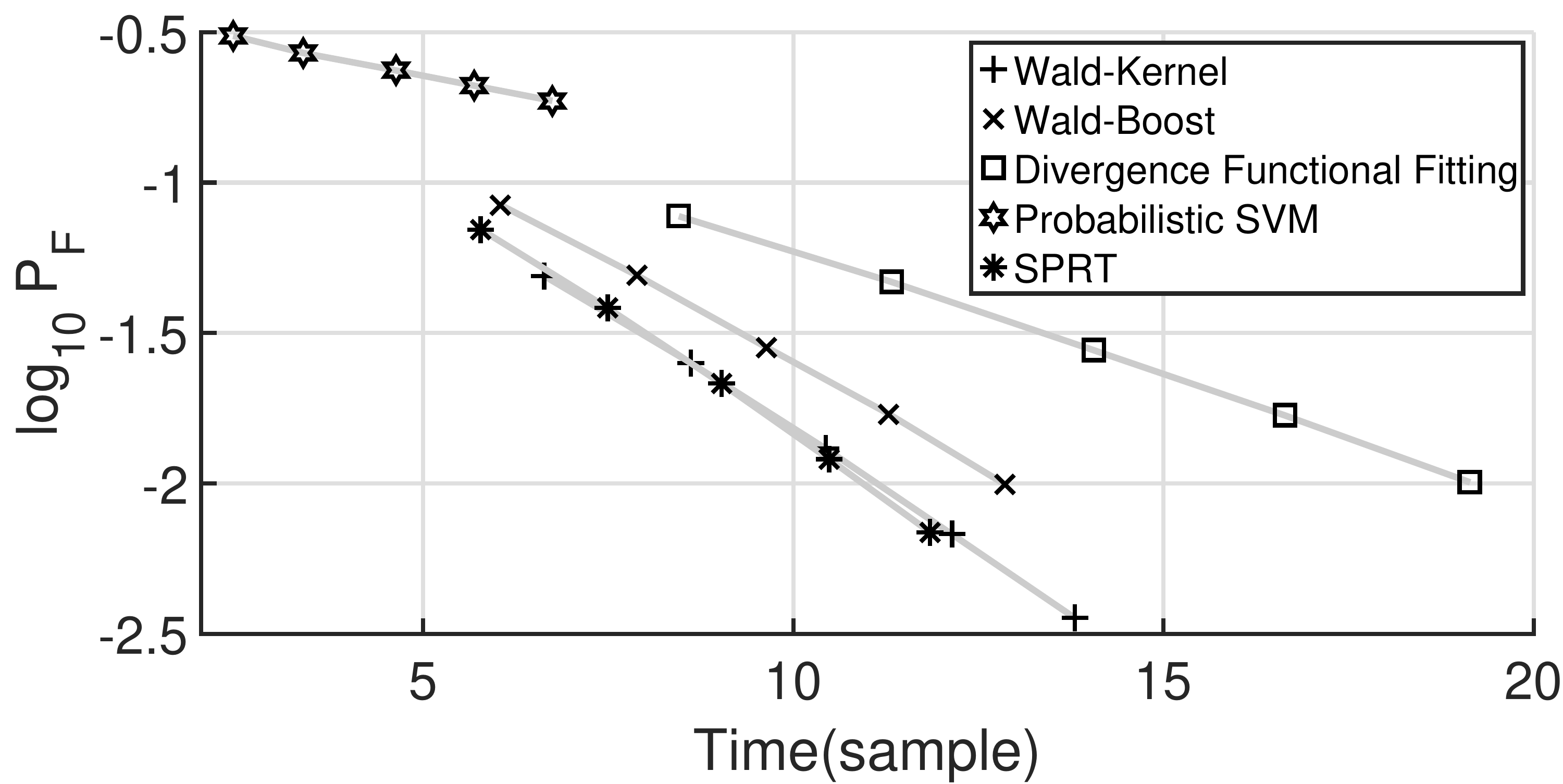}%
\caption{Probability of false alarm \emph{vs.} testing time}%
\label{synthetic_pf_time}
\end{subfigure}\hfill
\begin{subfigure}[t]{0.33\linewidth}
\includegraphics[width=\textwidth]{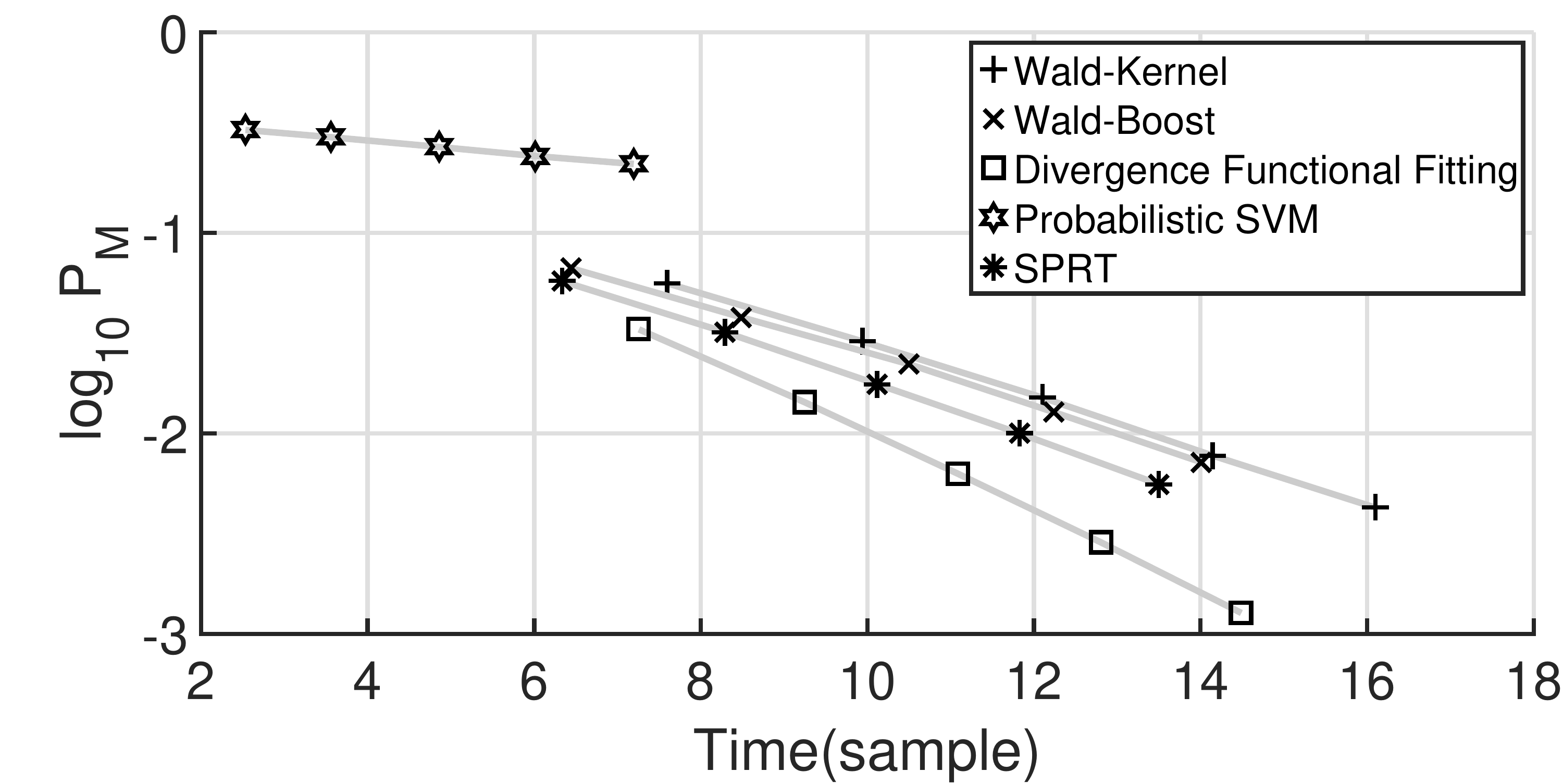}%
\caption{Probability of miss detection \emph{vs.} testing time}%
\label{synthetic_pm_time}
\end{subfigure}
\captionsetup{justification=centering}
\caption{Wald-Kernel \emph{vs.} other likelihood-ratio estimation approaches: probability of error \emph{vs.} testing time under each class}
\label{synthetic_err_time_performance}
\hrule
\end{figure*}

We plot the simulated performance curve in Figure~\ref{synthetic_err_time_performance}. As we may notice in Figure~\ref{synthetic_err_time}, Wald-Kernel achieve the best time-accuracy tradeoff among all the approaches. Wald-Boost and one-sided KL divergence estimator are slightly worse than Wald-Kernel in this example. Meanwhile, the Probabilistic SVM performs the worst in this example and fails to aggregate information over samples efficiently. Next, we may notice that the one-sided KL divergence estimation approach is completely biased towards $\mathrm{H}_1$. This could be identified by comparing Figure~\ref{synthetic_pf_time} with Figure~\ref{synthetic_pm_time}. Even though in Figure~\ref{synthetic_pm_time} the one-sided KL divergence estimation approach seems surprisingly outstanding (even better than SPRT), under the $\mathrm{H}_0$ case it performs awkwardly. And thus the overall performance of one-sided KL approach was not balanced correctly compared with Wald-Kernel which tailors the estimation procedure specifically for the sequential testing task. Similarly, Wald-Boost approach seems to outperform Wald-Kernel in the $\mathrm{H}_1$ case. But, overall the performance degrades when both classes are mixed due to its performance under $\mathrm{H}_0$ not being balanced correctly.

\begin{figure*}[h]
\centering
\includegraphics[width=\linewidth]{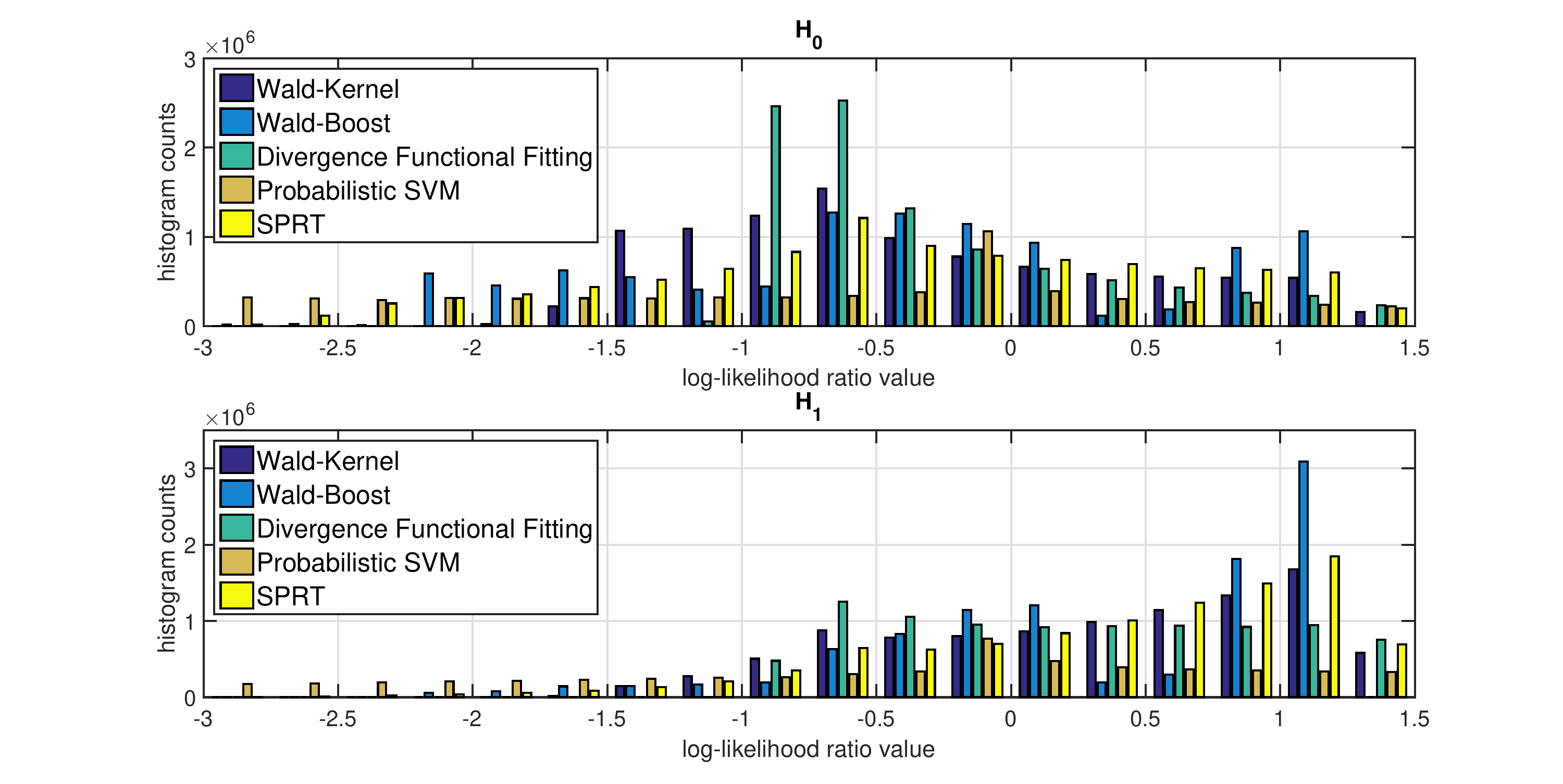}%
\captionsetup{justification=centering}
\caption{Wald-Kernel \emph{vs.} other likelihood-ratio estimation approaches: histogram of estimated log-likelihood ratio value under each class}%
\label{synthetic_histogram}
\hrule
\end{figure*}

To better analyze the likelihood ratio prediction performance and link it to the testing performance, we plotted the same empirical log-likelihood ratio values histogram as in the previous section in Figure~\ref{synthetic_histogram}. First, the Probabilistic SVM approach seems widely spread and provides much heavier tails than other methods. Thus, it explains why Probabilistic SVM terminates very fast yet not being successful in aggregating the class information. Second, one-sided KL approach did not predict any value smaller than $-1.25$ under both hypotheses. Thus, under $\mathrm{H}_0$ it terminates very slow and also become biased towards $\mathrm{H}_1$. Finally, for Wald-Boost, it predicts too many high values biased towards $\mathrm{H}_1$ side, causing it to become inaccurate in the $\mathrm{H}_0$ case. In conclusion, Wald-Kernel approach seems the most successful one in balancing the two classes. 

\subsection{Human Activity Recognition Using Mobile Health Sensor}
Next, we perform the sequential testing task using real world example. We consider classifying human activity using wearable mobile health sensor. The dataset we used is the UCI Human Activity Recognition Using Smartphones Data Set~\cite{anguita2013public}. The dataset contains continuously measured accelerometer and gyroscope data sampled at 50Hz when the subjects are performing 6 types of activity: walking, walking upstairs, walking downstairs, sitting, standing and laying. Later, a 561 dimensional feature vector is generated from the time domain 6-axis measurements. In this section, we choose the binary detection task for testing walking upstairs ($\mathrm{H}_0$) against walking downstairs ($\mathrm{H}_1$). The features we selected are feature1-3 concatenated with feature121-123, which are time domain mean value of the accelerometer and gyroscope measurements. Here, the choice of feature is not targeted to provide optimal discriminant between the two classes but to provide a reasonable amount of information for performing the sequential tests. \footnote[1]{In comparison, some feature from the dataset is almost linearly separable and super informative. We avoid choosing such feature because otherwise the test terminates in a single time slot and almost at zero probability of error. This would make the sequential testing trivial. Also it becomes simply a single-shot classifier comparison, which is not of our interest in this paper.} We extract the 6 dimensional feature from their training and testing file to form a 1073/471-by-6 training/testing set for $\mathrm{H}_0$ and a 986/420-by-6 training/testing set for $\mathrm{H}_1$.

To train the Wald-Kernel classifier, we selected 200 kernel centers using k-means clustering algorithm from the class mixed training set. The same kernel centers are used in the one-sided KL approach. For both approaches, we select Gaussian kernel with kernel width chosen by 50/50 cross validation from the candidates pool of the $\{1,5,10,25,50\}$th percentiles of the pairwise Euclidean distance between the kernel centers to all training samples. Next, for Wald-Boost, we choose the same parameter as previous synthetic example. We use 200 weak-learners with each weak-learner being a two-layer decision tree. For, Probabilistic SVM, the same 10-fold cross validation procedure is applied to train the SVM and select the posterior probability transformation coefficients.

\begin{figure}[h]
\centering
\includegraphics[width=\linewidth]{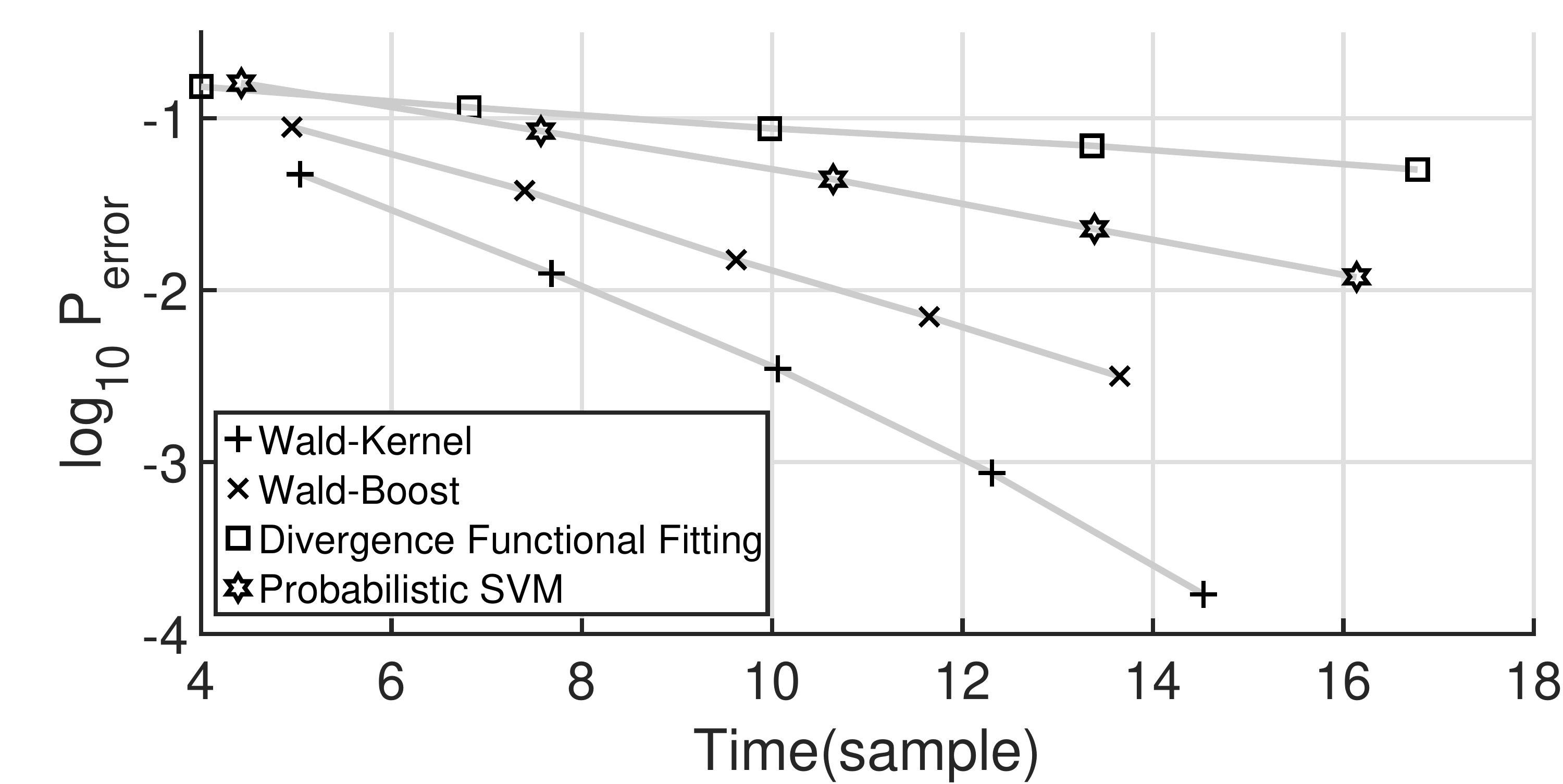}
\captionsetup{justification=centering}
\caption{Human activity recognition using mobile health sensor: walking upstairs \emph{vs.} walking downstairs}
\label{activity_performance}
\end{figure}

The testing phase in this example is slightly different compared with the synthetic example. Since the total number of training data is limited to 471/420 each class, we cannot arbitrarily obtain infinite amount of testing samples. Thus, we uniformly sample from the testing sample pool with replacement to perform the sequential testing task. Unfortunately, the log-likelihood ratio value reported by different approaches are highly diverse in terms of the concentration range in this example. We scale the output log-likelihood value to ensure the sequential testing performance curve appear roughly at the same region for visualization simplicity. \footnote[2]{Scaling the log-likelihood value does not affect the offset and slope of the time-log-accuracy tradeoff line but only affect where a particular termination boundary pair appears in that line. For example, in the synthetic case the Probabilistic SVM is not spanning the figure's entire zone. But if we scale its output down by a factor of 3, the offset and slope will not be affected, but could potentially make it appears roughly at similar region as the other methods.} Specifically, for Wald-Boost we scale its output by 0.7. For one-sided KL approach, we scale by 1.8. And for Probabilistic SVM we scale by 0.6. Next, the time-accuracy performance plot is presented in Figure~\ref{activity_performance}. The results show that the Wald-Kernel approach out performs the other approaches by aggregating information over samples more efficiently. The resulting probability of error of Wald-Kernel approach decrease much faster than other approaches.

\subsection{Military Vehicle Recognition Using Doppler Radar}
Finally, we present an example of using Doppler radar images to classify military vehicles. The dataset we select is the Man-portable Surveillance and Target Acquisition Radar (MSTAR) dataset. We choose to detect BTR-70 ($\mathrm{H}_0$) which is a military transportation vehicle against T-72 ($\mathrm{H}_1$) which is a tank. We apply Locality Preserving Projection (LPP)\cite{he2004locality} to extract a 16-dimensional feature from the Doppler radar image.

\begin{figure}[h]
\centering
\includegraphics[width=\linewidth]{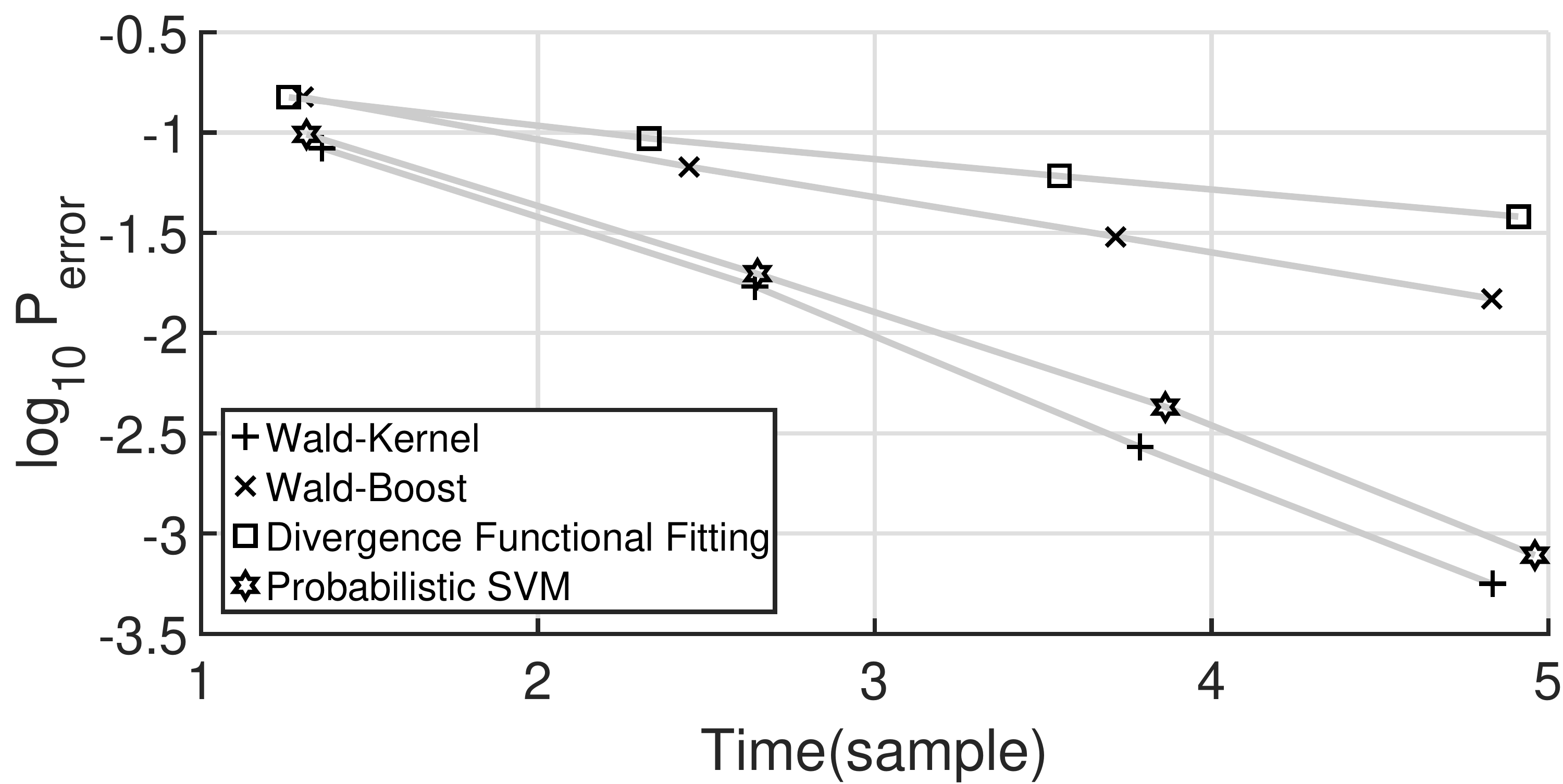}
\captionsetup{justification=centering}
\caption{Military targets recognition using Doppler radar image: BTR-70 \emph{vs.} T-72}
\label{mstar_performance}
\end{figure}

Here the dataset for the two classes is imbalanced in size. $\mathrm{H}_0$ has 1273 sample images in total while $\mathrm{H}_1$ has 429 sample images in total. We select the first 600 images from $\mathrm{H}_0$ and the first 200 images from $\mathrm{H}_1$ as the training set and use the remaining images for testing. For training phase, the identical choices of parameter as from previous example are applied in this example. And same uniform sampling with replacement procedure is used in testing phase. For scaling the output log-likelihood ratio value, we use 2.35 for one sided KL approach, 0.2 for Wald-Boost, and 0.45 for Probabilistic SVM. The time-accuracy performance plot is presented in Figure~\ref{mstar_performance}. Again, among all approaches, Wald-Kernel seems the optimal one in this example. Meanwhile, Probabilistic SVM approach is very competitive in this example.

We summarize the experimental results from all examples as follows. First, Wald-Kernel QC version is a decent approximation for providing a computationally efficient alternative to the Wald-Kernel solution. Second, Wald-Kernel algorithm is more successful compared with other approaches by optimizing the classifier specifically for the sequential testing objective. The benefits of directly tailoring the log-likelihood ratio estimator could be identified when testing samples are class mixed with the correct prior probability. Finally, for the other approaches compared in this section, they tend to be unstable across different examples. It seems difficult to predict their performance.

\section{Conclusion}\label{sec:conc}
In this work, we propose Wald-Kernel method for learning binary sequential detectors. The proposed approach combines standard Wald's SPRT with variational optimization approach for constructing likelihood ratio estimator. Guarantees for the constructed likelihood ratio estimator to achieve standard SPRT performance asymptotically are provided. Future directions of the work include considering multiclass extension as well as multimodal sensor fusion.


%

\appendices

\section{Proof of Theorem~\ref{thm2}}\label{proof_thm2}
\begin{proof}
Recall that the expected number of sample in the standard SPRT is given in \eqref{sprt_samp_cost}. Since the terms inside the bracket are constant after fixing the error rate, we define the following two constants for simplicity:
\begin{align*}
\omega_0 = [\mathrm{P}_\mathrm{F} \log \frac{\mathrm{P}_\mathrm{F}}{1-\mathrm{P}_\mathrm{M}} + (1-\mathrm{P}_F) \log \frac{1-\mathrm{P}_\mathrm{F}}{\mathrm{P}_\mathrm{F}}]
\end{align*}
and
\begin{align*}
\omega_1 = [\mathrm{P}_\mathrm{M} \log \frac{\mathrm{P}_\mathrm{M}}{1-\mathrm{P}_\mathrm{F}} + (1-\mathrm{P}_\mathrm{M}) \log \frac{1-\mathrm{P}_\mathrm{M}}{\mathrm{P}_\mathrm{F}}]
\end{align*}
The standard SPRT sampling cost then can be written as:
\begin{equation}\label{cost_sprt_sampling_appendix}
\begin{split}
\mathrm{C} & \approx \frac{\pi_0 \omega_0}{\mathrm{D}_{01}} + \frac{\pi_1 \omega_1}{\mathrm{D}_{10}}\\
& = \frac{\pi_0 \omega_0}{-\int  \log r \mathrm{d} P_0} + \frac{\pi_1 \omega_1}{\int  \log r \mathrm{d} P_1}
\end{split}
\end{equation}
Next the cost objective can be upper bounded using the convex conjugate formula for $- \log(\cdot)$ function. For an extended real functional:
\begin{equation}
f: \mathbf{x} \to \mathbb {R} \cup \{+\infty \}
\end{equation}
the convex conjugate
\begin{equation}
f^{\star }: \mathbf{x}^{*}\to \mathbb {R} \cup \{+\infty \}
\end{equation}
is defined by the maximization problem:
\begin{equation}
f^{\star} (\mathbf{x}^{\star}) = \sup_{\mathbf{x}} \{ \langle \mathbf{x} \,, \mathbf{x}^{\star} \rangle - f(\mathbf{x})\}.
\end{equation}
Hence, for
\begin{equation}
f(x) = 
\begin{cases}
-\log (x),\quad & \text{for } x>0;\\
+\infty,\quad & \text{for } x\leq0.
\end{cases}
\end{equation}
the convex conjugate dual is
\begin{equation}
f^{\star}(x^{\star}) = 
\begin{cases}
-1-\log (x^{\star}),\quad & \text{for } x^{\star}<0;\\
+\infty,\quad & \text{for } x^{\star}\geq0.
\end{cases}
\end{equation}
Therefore, we may upper bound~\eqref{cost_sprt_sampling_appendix} as:
\begin{equation}
\begin{split}
&\quad \frac{\pi_0 \omega_0}{- \int  \log r \mathrm{d} P_0} + \frac{\pi_1 \omega_1}{\int  \log r \mathrm{d} P_1}\\
\leq & \quad   \frac{\pi_0  \omega_0}{\sup_g \int (g \cdot r + \log(-g) + 1) \mathrm{d} P_0}\\
+ & \quad  \frac{\pi_1 \omega_1}{\sup_f \int (f \cdot r^{-1} + \log(-f) + 1) \mathrm{d} P_1}\\
\leq & \quad  \inf_{f} \frac{\pi_0  \omega_0}{\int \frac{1}{f} \mathrm{d}P_1 - \log(-f) \mathrm{d} P_0 + \mathrm{d}P_0}\\
+ & \quad  \frac{\pi_1 \omega_1}{\int f \mathrm{d}P_0 + \log(-f) \mathrm{d}P_1 + \mathrm{d}P_1}
\end{split}
\end{equation}
Together with the two constraints~\eqref{normalization_constraint} we obtain the variational bound given in \eqref{objective} by setting
\begin{equation}
\int \frac{1}{f} \mathrm{d}P_1 + \mathrm{d}P_0 = 0 \quad \text{and} \quad \int f \mathrm{d}P_0 + \mathrm{d}P_1 = 0.
\end{equation}
The resulting estimator $f^{\star}$ is linked to $\hat{r}$ through:
\begin{equation}
\hat{r} = -{f^{\star}}.
\end{equation}
\end{proof}

\section{Proof of Theorem~\ref{thm_consistency1}}\label{proof_thm_consistency1}
We first introduce the following Theorem from empirical process~\cite[Theorem 3.7]{van2000applications}.
\begin{theorem1}[Uniform Law of Large Numbers under Entropy Condition]
Assume the envelop condition $\int G \mathrm{d} P < \infty$ and suppose that
\begin{equation*}
\frac{1}{m} \mathcal{H}_1 (\delta,\mathcal{G},P_{m}) \xrightarrow{P} 0, \quad \forall \delta>0.
\end{equation*}
Then $\mathcal{G}$ satisfies ULLN.
\end{theorem1}
Next we prove Theorem~\ref{thm_consistency1}.
\begin{proof}
We evaluate the difference between the two terms in the first result of Theorem~\ref{thm_consistency1}:
\begin{align*}
& \int \log \hat{r} \mathrm{d}P_{1m} - \mathrm{E}[\log r |\mathrm{H}_1]\\
= & \int \log \hat{r} \mathrm{d}P_{1m} - \log r \mathrm{d}P_1\\
= & \int \underbrace{\log \hat{r} \mathrm{d}(P_{1m} - P_{1})}_{\mathrm{I}} + \underbrace{(\log \hat{r} - \log r) \mathrm{d}P_1}_{\mathrm{II}}
\end{align*}
By assumption $\int_{\mathbf{x}\in\mathcal{X}} \sup_{\hat{r}\in\mathcal{R}} |\log\hat{r}(\mathbf{x})| p_1(\mathbf{x}) \mathrm{d} \mathbf{x} < \infty$ and $\frac{1}{m} \mathcal{H}_1 (\delta,\log\mathcal{R},P_{1m}) \xrightarrow{P_1} 0$. Thus term $\mathrm{I} \xrightarrow{a.s.} 0$. Next, Assumption~\ref{asmp_modeling} implies $\mathrm{II} \xrightarrow{a.s.} 0$. Finally, we have
\begin{equation*}
\int \log \hat{r} \mathrm{d}P_{1m} \xrightarrow{a.s.} \mathrm{E}[\log r |\mathrm{H}_1].
\end{equation*}
And similarly, one can prove
\begin{equation*}
\int -\log\hat{r} \mathrm{d}P_{0m} \xrightarrow{a.s.} \mathrm{E}[-\log r |\mathrm{H}_0].
\end{equation*}
\end{proof}

\section{Proof of Theorem~\ref{thm_consistency2}}\label{proof_thm_consistency2}
We first provide a few important results that are required in evaluating the performance of a SPRT using learned likelihood ratio function $\hat{r}(\mathbf{x})$.
\begin{theorem1}\label{thm1a}
Consider the case that a SPRT is performed using an estimated likelihood ratio $\hat{r}(\cdot)$ instead of the ground truth $r(\cdot)$. If the estimated likelihood ratio function $\hat{r}(\cdot)$ is not constant and is normalized as:
\begin{equation}\label{normalization_constraint}
\int \hat{r} \mathrm{d}P_{0m} = 1 \quad \text{and} \quad \int \hat{r}^{-1} \mathrm{d}P_{1m} = 1
\end{equation}
then for fixed lower and upper thresholds $a$ and $b$ on log-likelihood ratio domain, probability of false alarm and miss detection of terminal decisions can be approximated by:
\begin{equation}\label{error_rate1}
\mathrm{P}_\mathrm{F}(\hat{r}) \approx \frac{1-e^{\tau_0 a}}{e^{\tau_0 b}-e^{\tau_0 a}}
\end{equation}
and
\begin{equation}\label{error_rate2}
\mathrm{P}_\mathrm{M}(\hat{r}) \approx \frac{e^{\tau_1 a}(e^{\tau_1 b}-1)}{e^{\tau_1 b}-e^{\tau_1 a}}
\end{equation}
where $\tau_0$ and $\tau_1$ are two positive constants satisfying:
\begin{align*}
\mathrm{E}[\hat{r}^{\tau_0}|\mathrm{H}_0] = 1\\
\mathrm{E}[\hat{r}^{-\tau_1}|\mathrm{H}_1] = 1
\end{align*}
\end{theorem1}
The existence of $\tau_0$ and $\tau_1$ is proved in~\cite[Theorem~3.1-(2)]{zou2010cooperative}. Next, to prove Theorem~\ref{thm1a}, the following Lemma is required.
\begin{theorem2}\label{lem1}
Let $\hat{z}_i = \log \hat{r}(\mathbf{x}_i)$, $\hat{\Lambda}_n = \sum_{k=1}^n \hat{z}_k$ and $\hat{G}(u) = \mathrm{E}[e^{u \hat{z}}]$, under both hypotheses the following process is a Martingale:
\begin{equation}\label{martingale}
\hat{\mathcal{M}}_n = \frac{e^{u \hat{\Lambda}_n}}{\hat{G}(u)^n}
\end{equation}
\end{theorem2}
\begin{proof}
First of all, it is easy to check $\hat{\mathcal{M}}_0=1$. Next, one can verify that:
\begin{align*}
\mathrm{E}[\hat{\mathcal{M}}_{n+1}|\hat{\mathcal{M}}_{n},0\leq k \leq n] & = \mathrm{E}[\frac{e^{u \hat{\Lambda}_{n+1}}}{\hat{G}(u)^{n+1}}|\hat{\mathcal{M}}_k,0 \leq k \leq n]\\
& = \mathrm{E}[\frac{e^{u \hat{z}_{n+1}}}{\hat{G}(u)} \cdot \hat{\mathcal{M}}_n | \hat{\mathcal{M}}_k, 0 \leq k \leq n]\\
& = \hat{\mathcal{M}}_n \cdot \frac{\mathrm{E}[e^{u \hat{z}_{n+1}}]}{\hat{G}(u)}\\
& = \hat{\mathcal{M}}_n
\end{align*}
And since $\hat{\mathcal{M}}_n$ are all positive valued, we have:
\begin{align*}
\mathrm{E}[|\hat{\mathcal{M}}_n|] = \mathrm{E}[\hat{\mathcal{M}}_n] = \mathrm{E}[\hat{\mathcal{M}}_0] = 1
\end{align*}
Thus we proved the process $\hat{\mathcal{M}}_n$ satisfies the two properties to be a Martingale.
\end{proof}
Next we prove Theorem~\ref{thm1a}.
\begin{proof}
Define $\hat{G}_0(u) = \mathrm{E}[e^{u \hat{z}}|\mathrm{H}_0]$ and $\hat{G}_1(u) = \mathrm{E}[e^{-u  \hat{z}}|\mathrm{H}_1]$. The special case of $\hat{r} \equiv 1$ satisfies both constraints, but when $\hat{r} \equiv 1$ the test never stops. Other than that, there is no constantly valued $\hat{r}(\cdot)$ satisfies both constraints. When $\hat{r}$ is not constantly $1$, the test will stop at finite time. Let $N$ be the random stopping time, then we have the two types of error when the test stops:
\begin{align*}
\mathrm{P}_\mathrm{F}(\hat{r}) = \Pr \{\hat{\Lambda}_N \geq b|\mathrm{H}_0\} \quad \text{and} \quad \mathrm{P}_\mathrm{M}(\hat{r}) = \Pr \{\hat{\Lambda}_N \leq a|\mathrm{H}_1\}
\end{align*}
For the special case of $u=0$, $u=-1$ and $u=1$:
\begin{align*}
\hat{G}_0(0) = 1, \quad \hat{G}_1(0) = 1
\end{align*}
and
\begin{align*}
\hat{G}_0(\tau_0) = \int \hat{r}^{\tau_0} P_0 = 1, \quad  \hat{G}_1(\tau_1) = \int \hat{r}^{-\tau_1} P_1 = 1
\end{align*}
Now one can evaluate the expected value of the Martingale $\hat{M}_N$ at the stopping time $N$ under $\mathrm{H}_0$ with $u=1$, which gives:
\begin{align*}
& \mathrm{E}[\frac{e^{\tau_0 \hat{\Lambda}_N}}{\hat{G}_0(\tau_0)^N} | \mathrm{H}_0] = 1\\
\iff &  \mathrm{E}[e^{\tau_0 \hat{\Lambda}_N}|\mathrm{H}_0] = 1\\
\iff & \mathrm{P}_\mathrm{F}(\hat{r}) B^{\tau_0} + (1-\mathrm{P}_\mathrm{F}(\hat{r})) A^{\tau_0} \approx 1\\
\iff & \mathrm{P}_\mathrm{F}(\hat{r}) \approx \frac{1 - A^{\tau_0}}{B^{\tau_0} - A^{\tau_0}}
\end{align*}
where $A = e^a$ and $B = e^b$. The third approximation is due to Assumption~\ref{asmp1}. Similarly, one can get:
\begin{align*}
\mathrm{P}_\mathrm{M}(\hat{r}) \approx \frac{A^{\tau_1}(B^{\tau_1}-1)}{B^{\tau_1}-A^{\tau_1}}.
\end{align*}
\end{proof}

In addition to the changes in termination error, one can also evaluate the changes in termination time.

\begin{theorem1}\label{thm1b}
Assume a likelihood ratio estimator $\hat{r}$ satisfying Theorem~\ref{thm2} is constructed to perform SPRT task. The expected stopping time under each hypothesis could be evaluated as:
\begin{equation}\label{time_H0}
\mathrm{N}_0(\hat{r}) \approx \frac{b-a + \tau_0 a e^{\tau_0 b} - \tau_0 b e^{\tau_0 a}}{e^{\tau_0 b}-e^{\tau_0 a}} \cdot \frac{1}{\mathrm{E}[\log \hat{r} | \mathrm{H}_0]}
\end{equation}
and
\begin{equation}\label{time_H1}
\mathrm{N}_1(\hat{r}) \approx \frac{a-b  + \tau_1 b e^{- \tau_1 a}  - \tau_1 a e^{-\tau_1 b}}{e^{-\tau_1 a} - e^{-\tau_1 b}} \cdot \frac{1}{\mathrm{E}[\log \hat{r} |\mathrm{H}_1]}
\end{equation}
\end{theorem1}
\begin{proof}
We review Wald's equation that is required in this proof.
\begin{identity}[Wald's Equation]
Let $\{x_1,x_2,\cdots,x_N\}$ be a sequence of real valued i.i.d. random variables, and $N$ be a nonnegative random integer independent of the sequence. If both expectation $\mathrm{E}[N]$ and $\mathrm{E}[x]$ are finite, then:
\begin{equation}
\mathrm{E}[x_1+x_2+\cdots+x_N] = \mathrm{E}[N]\mathrm{E}[x]
\end{equation}
\end{identity}

We first verify the two conditions for Wald's equation are satisfied when $\log\hat{r}$ is used in the accumulation process. First, the non-constant condition guarantees that the two hypotheses are distinguishable:
\begin{equation*}
\mathrm{E}[\log \hat{r}|\mathrm{H}_1] > 0, \quad \mathrm{E}[-\log \hat{r}|\mathrm{H}_0] > 0
\end{equation*}
Next, from Assumption~\ref{asmp1} we have:
\begin{equation*}
\begin{split}
\mathrm{E}[\log \hat{r} |\mathrm{H}_1] \leq & \mathrm{E}[\log r |\mathrm{H}_1]\\
\leq & \mathrm{E}[(b \lor -a)|\mathrm{H}_1]\\
< & +\infty
\end{split}
\end{equation*}
And similarly:
\begin{equation*}
\mathrm{E}[-\log \hat{r} |\mathrm{H}_0] < +\infty
\end{equation*}
Finally, the stopping time is guaranteed to be finite by~\cite[Lemma~1]{wald1944cumulative}. Therefore, Wald's equation implies:
\begin{equation*}
\begin{split}
\mathrm{N}_0(\hat{r}) = & \frac{\mathrm{E}[\sum_{t=1}^{T_{stop}}\log \hat{r}(\mathbf{x}_t)|\mathrm{H}_0]}{\mathrm{E}[\log \hat{r}(\mathbf{x})|\mathrm{H}_0]}\\
\approx & \frac{b-a + \tau_0 a e^{\tau_0 b} - \tau_0 b e^{\tau_0 a}}{e^{\tau_0 b}-e^{\tau_0 a}}\cdot\frac{1}{\mathrm{E}[\log \hat{r}|\mathrm{H}_0]}
\end{split}
\end{equation*}
where $\mathrm{E}[\sum_{t=1}^{T_{stop}}\log \hat{r}(\mathbf{x}_t)|\mathrm{H}_0] \approx \hat{\mathrm{P}}_{\mathrm{F}} b + (1-\hat{\mathrm{P}}_{\mathrm{F}})a$. And similarly, we can obtain:
\begin{equation*}
\mathrm{N}_1(\hat{r}) \approx \frac{a-b  + \tau_1 b e^{-\tau_1 a}  - \tau_1 a e^{-\tau_1 b}}{e^{-\tau_1 a} - e^{-\tau_1 b}} \cdot \frac{1}{\mathrm{E}[\log \hat{r} |\mathrm{H}_1]}.
\end{equation*}
\end{proof}

Next, we prove Theorem~\ref{thm_consistency2}.
\begin{proof}
To prove the consistency result in terms of probability of error and expected time to terminate, it is equivalent to prove:
\begin{equation}
\lim_{m \to \infty} \tau_0 = 1
\end{equation}
and
\begin{equation}
\lim_{m \to \infty} \tau_1 = 1
\end{equation}
From Theorem~\ref{thm_consistency1}, we have:
\begin{equation}
\lim_{m \to \infty} \int \hat{r} \mathrm{d}P_{0m} = \int r \mathrm{d}P_{0} = 1
\end{equation}
Meanwhile, the following function:
\begin{equation}
\hat{G}_{0m}(\tau_0) = \lim_{m \to \infty} \int e^{\tau_0 \log \hat{r}} P_{0m}
\end{equation}
is convex in $\tau_0$. And if we allow $\tau_0 = 0$, we have:
\begin{equation}
\lim_{m \to \infty} \left. \int \hat{r}^{\tau_0} \mathrm{d}P_{0m} \right\rvert_{\tau_0 = 0} = 1
\end{equation}
Next, we show that there does not exist a positive $\tau_0 \neq 1$ such that:
\begin{equation}\label{tau_condition}
\lim_{m \to \infty} \int \hat{r}^{\tau_0} \mathrm{d}P_{0m}  = 1
\end{equation}
This could be proved by contradiction. Suppose there exist a $0<\tau_0^\prime<1$ satisfying~\eqref{tau_condition}. Since the function $\hat{G}_{0m}(\tau_0)$ is convex in $\tau_0$, we have:
\begin{equation}
\tau_0^\prime \hat{G}_{0m}(1) + (1-\tau_0^\prime) \hat{G}_{0m}(0) > \hat{G}_{0m} (\tau_0^\prime)
\end{equation}
which contradicts with the fact $\hat{G}_{0m}(1) = \hat{G}_{0m}(0) = \hat{G}_{0m}(\tau_0^\prime) = 1$. And similarly, one can prove that there is no $\tau_0 > 1$ satisfying~\eqref{tau_condition}. Next, a similar proof could be constructed to guarantee $\lim_{m \to \infty} \tau_1 = 1$, completing our proof.
\end{proof}



\ifCLASSOPTIONcaptionsoff
  \newpage
\fi



%

\bibliographystyle{IEEEtran}
\bibliography{wald_kernel_ref}

%

%







\end{document}